%% file: main.tex
\documentclass[lettersize,journal]{IEEEtran}

\usepackage{amssymb}
\usepackage{amsthm}
\usepackage{amsmath}
\usepackage[c2]{optidef}
\usepackage[table]{xcolor}
\usepackage{subcaption}

\def\BibTeX{{\rm B\kern-.05em{\sc i\kern-.025em b}\kern-.08em
    T\kern-.1667em\lower.7ex\hbox{E}\kern-.125emX}}
\usepackage{balance}

\usepackage{widetext}
\usepackage{mathtools}
\usepackage[thinc]{esdiff}

\newcommand{\s}[1]{\prescript{*}{}{#1}}

\newcommand{\mn}{\mathbb{N}}

\newcommand{\mr}{\mathbb{R}}

\newcommand{\me}{\mathbb{E}}

\newcommand{\um}{\scalebox{0.5}[1.0]{\( - \)}}

\newtheorem{theorem}{Theorem}
\newtheorem{proposition}{Proposition}

\newtheorem{corollary}{Corollary}
\newtheorem*{corollary*}{Corollary}
\newtheorem*{theorem*}{Theorem}
\newtheorem*{definition*}{Definition}

\usepackage{balance}

\begin{document}

\title{Informed Deep Hierarchical Classification: a Non-Standard Analysis Inspired Approach}

 \author{Lorenzo Fiaschi and Marco Cococcioni
\thanks{Lorenzo Fiaschi and Marco Cococcioni are with the
Department of Information Engineering, University of Pisa, Pisa 56122,
Italy (e-mail: lorenzo.fiaschi@ing.unipi.it; marco.cococcioni@unipi.it)\\No conflict of interest to disclose.\\IEEE is credited as copyright holder of the present version of the manuscript.}}

\markboth{IEEE TRANSACTIONS ON NEURAL NETWORKS AND LEARNING SYSTEMS,,~Vol.~XX, No.~X, October~2024}%
{How to Use the IEEEtran \LaTeX \ Templates}

\maketitle

\begin{abstract}
This work proposes a novel approach to the deep hierarchical classification task, i.e., the problem of classifying data according to multiple labels organized in a rigid parent-child structure.
It consists of a multi-output deep neural network equipped with specific projection operators placed before each output layer.
The design of such an architecture, called lexicographic hybrid deep neural network (LH-DNN), has been possible by combining tools from different and quite distant research fields: lexicographic multi-objective optimization, non-standard analysis, and deep learning.
To assess the efficacy of the approach, the resulting network is compared against the B-CNN, a convolutional neural network tailored for hierarchical classification tasks, on the CIFAR10, CIFAR100  (where it has been originally proposed), and Fashion-MNIST benchmarks.
Evidence states that an LH-DNN can achieve comparable if not superior performance, especially in learning the hierarchical relations, in the face of a drastic reduction of the learning parameters, training epochs, and computational time, without the need for ad-hoc loss functions weighting values.
\end{abstract}


\section{Introduction}
\label{sec:introduction}
Hierarchical classification (HC) is a challenging machine learning problem where data are supposed to be classified by assigning them a set of labels arranged according to a hierarchy.
According to the taxonomy proposed by \cite{HMC}, HC problems can be represented by a triple $\langle\Upsilon,\Psi,\Phi\rangle$ where $\Upsilon$ specifies the type of graph representing the hierarchy, $\Psi$ indicates whether a data instance is allowed to have class labels associated with
single or multiple paths in the class hierarchy, and $\Phi$ describes the label depth of the data instance.
The present work focuses on triplets of the form $\langle T,S,F \rangle$, that is HC problems where labels are organized according to a tree-like structure (T), each data point is associated with one single path through the hierarchy (S), and each path travels the full depth of the tree (F).
The use of more general HC configurations is currently under investigation.

HC naturally arises in many applications of interest as text categorization \cite{lewis2004rcv1,peng2018large,rousu2006kernel}, image recognition and segmentation \cite{imagenet,mendoncca2021intrusion,taoufiq2020hierarchynet}, medicine \cite{murtaza2020breast,xie2018neural}, biology \cite{dimitrovski2012hierarchical,mahmood2020automatic}, functional genomics \cite{nakano2018improving,sandaruwan2021improved,vens2008decision} and protein classification, to mention a few.
Indeed, the introduction of the hierarchical information during the training process proved to be an effective technique to obtain classifiers manifesting better performance, able to learn from less data, and with stronger guarantees of compliance with the background knowledge itself \cite{giunchiglia2022dlconst}.
As pointed out in \cite{giunchiglia2020coherent} and \cite{xue2008deep}, such an additional form of reasoning comes at the price of some non-trivial challenges: the intrinsic imbalance between parent and child classes, exponentially longer training time with respect to the number of classes and how they are distributed in the hierarchy, an exponential decrease of accuracy of the same classifier when trained on datasets with a moderate larger number of classes.
Thus, the problem of how to clearly embed and efficiently exploit the parent-child relations becomes crucial for the effectiveness and usability of an algorithm.

The present work proposes a lexicographic hybrid HC approach, called lexicographic hybrid DNN (LH-DNN), that manages per-level knowledge within a single, global classifier.
Its novelty is represented by the way in which the local information is processed, being it inspired by results coming from the domain of non-standard analysis (NSA), a branch of Mathematics dealing with infinite and infinitesimal numbers.
By reformulating the HC problem as a lexicographic multi-objective optimization problem (LMOP) and interpreting it according to the tools proper of NSA, the article shows how it is possible to design deep neural networks (DNNs) whose learning process is compliant with the hierarchical structure of the data.
In particular, Section \ref{sec:hybrid_networks} is in charge of presenting the theoretical foundation of the proposal and possible implementation of it, while Section \ref{sec:experiments} is meant to assess the efficacy of the novel approach.
The two the sections are anticipated by preparatory ones: Section \ref{sec:nsa_and_at} provides the minimally needed background on NSA, while Section \ref{sec:hc_and_bcnn} provides a more detailed overview of the HC problem and recently developed algorithms, and presents the branching networks proposed in \cite{zhu2017b}, which will constitute the touchstone to assess the effectiveness of the proposal.
The choice of such networks for the experimental comparison is motivated for two main reasons: i) they provide a clear and fair benchmark to measure the impact of LH-DNNs on HC problems without external or incompatibility factors that may alter the study; ii) they notably influenced the recent research, as the literature presents numerous works based or heavily inspired by them \cite{li2023semantic,seo2019hierarchical,su2021hierarchical,verma2020yoga,wehrmann2018hierarchical}.
Finally, Section \ref{sec:conclusions} provides the conclusions.

\section{Non-standard analysis and the Alpha theory}
\label{sec:nsa_and_at}
NSA is a branch of mathematical logic aiming at developing models of analysis where infinite and infinitesimal numbers are allowed along with the more common finite, real ones.
Anticipated by numerous debates in ancient Greece and pioneered by the infinitesimal calculus of Leibniz and Newton, the first rigorous non-standard model of analysis saw the light of day in the `60s with the book ``Non-standard Analysis'' by A. Robinson \cite{robinson2016non}.
From that moment on, several alternatives have been proposed, each with its own upsides and downsides.
Among them, one relevant to the present work is the ``Alpha theory'' \cite{BDNbook}, which introduces a non-standard model of analysis through a very lightweight and logic-free axiomatization.
There are two reasons for which such a non-standard model of analysis has been chosen for the present work.
The first one is that Alpha theory is a technology ready to be used in practice by applied scientists like engineers, computer scientists, and economists, thanks to two factors: i) the simplicity of its presentation; the existence of a finite-length numerical binary encoding for its non-standard numbers \cite{Benci_J1}.
The second reason is the presence of a transfer principle, an aspect that will be discussed later in this section.

The use of numerical infinite and infinitesimal numbers in concrete applications is a quite recent trend that seems to be particularly fruitful, and among the research fields positively affected by such a novelty there is multi-objective optimization.
Non-finite numbers-based approaches have been adopted for linear programming\cite{CococcioniEtAlAMC2018,Fiaschi_J4}, mixed-integer linear programming \cite{Cococcioni_20,lexico-cutting_23}, quadratic programming \cite{astorino_20,DeLeone_svm,na_ipm}, and non-linear optimization \cite{DeLeone,Lai_et_al_PC,Lai_et_al_PL}, to mention a few.

The name Alpha Theory stems from the specific infinite value $\alpha\in\me\supset\mr$ ($\me$ is also indicated by $\s{\mr}$), whose existence is postulated within the theory itself.
A detailed presentation of the non-standard model is omitted for the sake of brevity, pointing the interested reader to \cite{BDNbook}.
What follows is instead an overview of the main concepts of the theory preparatory for the remaining of the work.

\begin{theorem*}
    $\me\supset\mr$ is a field, $\alpha\in\me$.
\end{theorem*}

\noindent The theorem states that $\me$ contains other numbers than the real ones, e.g., $\alpha$ and $\eta\coloneqq\alpha^{\um1}$.
Since $\mr$ contains all and only the finite numbers, the remaining numbers must be infinite and infinitesimal.
The next three definitions formally state when a number can be addressed as finite, infinite, or infinitesimal.

\begin{definition*}[Infinite number]
    $x\in\me$ is infinite $\xLeftrightarrow{def} \nexists\,k\in\mn$ such that $k>|x|$.
\end{definition*}

\begin{definition*}[Finite number]
    $x\in\me$ is finite $\xLeftrightarrow{def} \exists\,k\in\mn$ such that $k>|x|>\frac{1}{k}$.
\end{definition*}

\begin{definition*}[Infinitesimal number]
    $x\in\me$ is infinitesimal $\xLeftrightarrow{def} \nexists\,k\in\mn$ such that $|x|>\frac{1}{k}$.
\end{definition*}

The next theorem guarantees that any real function can be smoothly extended to values in $\me$.

\begin{theorem}
    Any function $f\colon\mr\rightarrow\mr$ can be coherently and uniquely extended to a function $\s{f}\colon\me\rightarrow\me$ such that $\s{f}(x) = f(x)\,\forall x\in\mr$.
    \label{theo:f_extension}
\end{theorem}

\noindent The fact that the previous theorem holds is necessary to state the next one without ambiguities.
Its role is to characterize the elements in $\me$, that is to explicit what are the non-standard numbers which can be used within the Alpha theory.

\begin{theorem*}
    The numbers in $\me$ are all and only the values at the point $\alpha$ of all the real functions extended to $\me$.
\end{theorem*}

\noindent With the purpose of an example and according to the last theorem, the following values belong to $\me$
\[
-\frac{\alpha}{3}+\eta^2, \qquad 3+2\eta, \qquad \frac{\alpha^7-\alpha^{\pi}+\eta^\alpha}{-3+5 e^\eta},
\]
and the following relations hold
\[
\alpha \cdot (2\alpha - 3) = 2\alpha^{2} - 3\alpha, \qquad 0 < \eta < \alpha^0 = 1 < \alpha < \alpha + 1,
\]
\[
\frac{8.06\alpha^2 +4.28 - 2.08\eta^3} {6.2\alpha^3-5.2} =  1.3 +0.4\eta^3.
\]

\begin{definition*}[Monosemium]
    Any value $\xi=r\alpha^k\in\me$, $r,k\in\mr$, is called monosemium
\end{definition*}

\begin{corollary*}
    $\xi\in\me \Longrightarrow\exists\{r_i\}_{i=1}^\infty$, $r_i\in\mr$ $\forall i$ and $\exists\{k_i\}_{i=1}^\infty$, $k_i\in\mr$ $\forall i$, $k_i > k_j$ $\forall i > j$, such that
    \[
    \xi = \sum_{i=1}^\infty r_i\alpha^{k_i}.
    \]
\end{corollary*}

The next theorem is fundamental in any non-standard model of analysis: informally, it guarantees that all the ``elementary properties'' that are true for certain mathematical objects in the standard model, say $\mr$, are also true for their non-standard counterparts in the non-standard model, say $\me$, and vice-versa.
Again, the reader is pointed to \cite{BDNbook} for a rigorous statement of the theorem and a proper definition of elementary property, which would have provided a tedious and unnecessary complex presentation of NSA.

\begin{theorem}[Transfer principle]
    An ``elementary property'' $\sigma$ is satisfied by mathematical objects $\varphi_1,\ldots,\varphi_k$ if and only if $\sigma$ is satisfied by their non-standard extensions $\s{\varphi_1},\ldots,\s{\varphi_k}$:
    \[
    \sigma(\varphi_1,\ldots,\varphi_k) \Longleftrightarrow \sigma(\s{\varphi_1},\ldots,\s{\varphi_k}).
    \] 
    \label{theo:transfer}
\end{theorem}


Finally, the authors of \cite{na_ipm} proved a theorem of equivalence between any standard lexicographic multi-objective optimization problem (LMOP) and a non-standard scalar program, which is of crucial importance for this work.
Informally, an LMOP is a problem where multiple functions are meant to be optimized but there exists a strict priority ordering among them.
Below, it is reported the formal definition.

\begin{definition*}[Lexicographic optimization problem]
    Let $f_1,\ldots,f_n$ be two real functions and there exists a strict priority ordering among them induced by the natural ordering. Then, the following sequence of programs constitutes a lexicographic optimization problem:\\
    \begin{minipage}{.4\linewidth}
        \begin{mini*}|s|
            {}{f_1(x)}{}{}
            \addConstraint{x\in\Omega}{}{}
        \end{mini*}
    \end{minipage}
    \begin{minipage}{.55\linewidth}
        \begin{mini*}|s|
            {}{f_i(x)}{}{}
            \addConstraint{f_j(x)}{=\Bar{f}_j}{\;\;\;\forall j=1,\ldots,i\um 1}
            \addConstraint{x\in\Omega}{}{}
        \end{mini*}
    \end{minipage}
    where $\Bar{f}_j$ is the optimal value found for $f_j$ in the previous optimization problems.
\end{definition*}

\begin{theorem}
    Consider an LMOP whose objective functions $f_1,\ldots,f_n$ are real functions and the priority is induced by the natural order.
    Then, there exists an equivalent scalar program over the same domain, whose objective function is non-standard and has the following form: 
    \begin{equation*}
    \vspace{-.3cm}
    f(x) = \beta_1 f_1(x) + \ldots + \beta_n f_n(x),
    \end{equation*}
    where $\beta_1,\ldots,\beta_n$ are non-standard values such that $\frac{\beta_{i+1}}{\beta_i}\approx0$ (i.e., is infinitesimal), $i=1,\ldots,n$-$1$.
    \label{theo:nsa_equivalence}
\end{theorem}

\noindent Before concluding, it is worth mentioning that where no ambiguities subsist, the prescript $\s{}$ is omitted for the sake of readability. 
An example is the expected value function in Equation \eqref{eq:lex_ns_hc}.

\section{Hierarchical Classification and Branch Neural Networks}
\label{sec:hc_and_bcnn}
According to \cite{HMC}, each HC algorithm can be classified according to a four-tuple $\langle\Upsilon,\Psi,\Phi,\Theta\rangle$, where the first three entries have the same meaning as before, while the fourth parameter, $\Theta$, indicates how the topological information of the class relationships is managed and used.
According to it, HC algorithms can be split into four families: \textit{local classifiers per node}, \textit{local classifiers per level}, \textit{local classifiers per parent node}, and \textit{global classifiers}.
In general, global approaches are usually cheaper than local ones, and they do not suffer from error-propagation issues at inference time, though they are less likely to capture local information from the hierarchy, risking underfitting.
Local approaches are much more computationally expensive since
they rely on a cascade of classifiers, but they are much more suitable for extracting information from regions of the class hierarchy, risking overfitting.

Local classifiers per node are HC algorithms constituted of multiple binary classifiers meant to discern whether or not a certain data point belongs to one specific class in the hierarchy.
An example is the model HMC-MLPN \cite{feng2018hierarchical}, implemented as a separate fully connected DNN for each label.
Since this approach scales poorly with respect to the number of labels and structurally suffers from class instances imbalance, it is rarely used in practice.

Local classifiers per parent node approaches train one classifier for each splitting node of the hierarchy, i.e., each parent node.
HC algorithms of this family, as \cite{kowsari2017hdltex} for texts or \cite{kulmanov2018deepgo} for gene ontologies, categorize data points feeding them to the various classifiers in cascade, starting from the root of the hierarchy and following the path induced by the predictions at each node.
Even if more common than the previous ones, local classifiers per parent node still struggle in scaling with respect to the number of labels, limiting their usability.

Local classifiers per level are HC algorithms constituted of a very limited number of classifiers: one for each level of the hierarchy.
The possibility to scale better with the number of labels and the absence of imbalance-inducing properties allowed this approach to become by far the most popular strategy to deal with HC problems locally.
Among the notable works, one may mention \cite{cerri2014hierarchical} and \cite{cerri2016reduction} for protein function prediction, and \cite{li2018deepre} and \cite{zou2019mldeepre} for enzyme multi-function prediction.

Finally, global classifier approaches deploy one single classifier able to capture the entire hierarchy at once.
To do so, several techniques are at disposal: injection of the hierarchical information directly in the network topology \cite{masera2019awx}, use of a constraint final layer \cite{giunchiglia2021multi}, leveraging capsule networks \cite{aly2019hierarchical}, even with an ad-hoc capsule-oriented loss function \cite{peng2019hierarchical}.
It is also possible to use postprocessing routines to enforce the hierarchy coherence \cite{obozinski2008consistent}.
Another frequent phenomenon is to combine global and per-level local techniques, implementing the so-called hybrid approaches such as \cite{li2023semantic,verma2020yoga,gao-2020-deep} and, more in general, the pioneering branching convolutional neural networks \cite{zhu2017b}, which had a significant number of follow-ups in the recent years.

\subsection{Branching Networks}
\begin{figure}[ht]
    \centering
    \includegraphics[width=.8\linewidth]{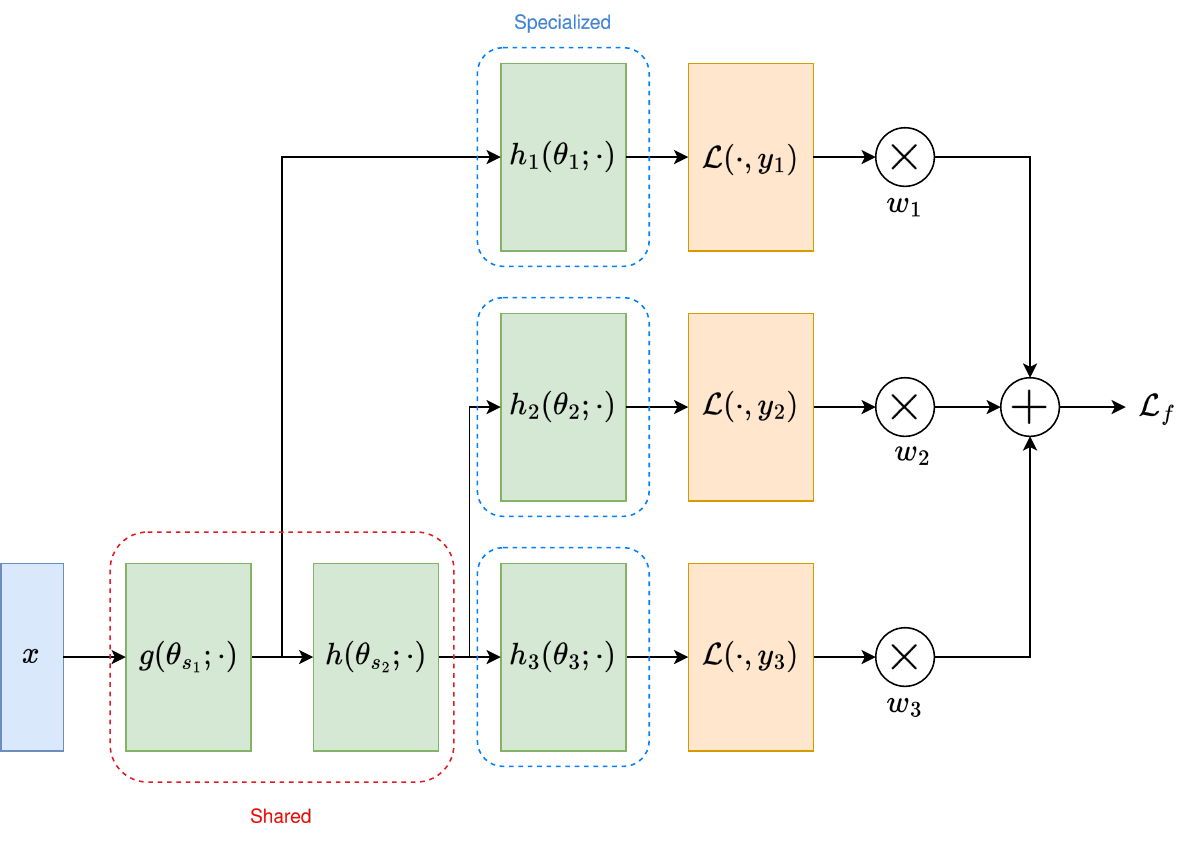}
    \caption{Example of a B-DNN for a three-level HC problem.}
    \label{fig:B-DNN}
\end{figure}

Branching deep neural networks (B-DNNs) are DNNs meant to output multiple predictions possessing a certain logical order, e.g., a priority among them.
The original and main application that saw their use concerns HC problems, where each prediction is dedicated to a specific hierarchy level.
To facilitate the presentation, Figure \ref{fig:B-DNN} reports an illustrative description of a B-DNN for a three-level HC problem.

B-DNNs consist of two main parts: the shared sub-network and the specialized sub-networks.
The first part is meant to extract features from data that will be used for the classification at all the hierarchy levels.
In particular, the coarser the categorization the smaller the portion of shared parameters it can train and the earlier they are in the feature extraction process.
With reference to Figure \ref{fig:B-DNN}, the shared parameters $\theta_{s_1}$ are in common to all the three branches, while $\theta_{s_2}$ only to the second and the third one.

The second part of the network is constituted by independent sub-networks, each specialized for a different level of the classification.
Such networks branch from the feature extraction flow at different points, assuming that coarser categorizations require less fine knowledge retrieval.
The errors made at each level of the hierarchy are gathered in one single value through a convex combination having time-dependent scalarization weights.
With reference to Figure \ref{fig:B-DNN}, the sub-network $h_i$ is specialized for the $i$-th level classification, and the error made by it is scaled by a factor $w_i$ before being summed with the other losses.

The reason for the time-varying scalarization is the will to mimic hierarchical learning during the network training.
The weights are initialized with the triple $(w_1,w_2,w_3) = [1,0,0]$, and smoothly shifted towards $[0,0,1]$ passing through a configuration where $w_2$ is significantly larger than the others.
In this way, the network starts learning the coarser classification task and gradually takes into consideration the other levels too, preserving their order in the hierarchy.
Moreover, the training of the feature extraction process manifests a similar behavior.
In the beginning, only the early shared weights ($s_1$ in Figure \ref{fig:B-DNN}) are optimized: the goal of such learning is to extract useful features for the coarser classification.
While the importance of lower hierarchy levels grows, also the other shared weights ($s_2$) start to be optimized.
Being after already trained parameters, they are likely modified to improve the already existing but coarser feature extraction.

\section{Lexicographic Hybrid DNNs: hierarchical classification from a non-standard perspective}
\label{sec:hybrid_networks}

\subsection{Hierarchical classification as a lexicographic problem}
The first step towards an NSA-inspired DNN requires the reformulation of an HC problem into an LMOP.
Generally speaking, an HC problem can be seen as the minimization problem in Equation \eqref{eq:hc_problem}
\begin{mini}|s|
    {\theta}{\mathcal{L}(f(\theta;X),Y,\Upsilon)}{}{\label{eq:hc_problem}}
\end{mini}
where a certain loss function $\mathcal{L}(\cdot)$ is meant to be minimized by adjusting the free parameters $\theta$ of a classification map $f$, e.g., a DNN, considering a dataset $X$, its set-labels $Y$, and the hierarchy structure $\Upsilon$. 
As anticipated, the work focuses on tree-like hierarchies, i.e., $\Upsilon = T$, and so each set-label possesses a well-defined structure $Y=(Y_1,\ldots,Y_n)$, $Y_1 = \{Y_1^{(x)}\}_{x\in X}$, where $Y_i^{(x)}$ indicates the label of $i$-th level of the hierarchy for the data point $x$, and such that there exists a parent-child relation between labels $Y_i^{(x)}$ and $Y_{i+1}^{(x)}$.

Embedding and leveraging the hierarchy structure $\Upsilon$ in the learning of the parameters $\theta$ is crucial for the design of a well-performing classifier.
Hybrid approaches combining local and global losses, e.g., B-DNNs seem to implicitly implement a parameter training that prioritizes the performance of the classifier on the coarser levels of the tree.
The main benefit of such a choice consists in the learning of inner representations of the data that are not only discriminant for the categorization at the higher levels of the hierarchy but also helpful for discerning at lower ones, as a sort of knowledge propagation typical of human reasoning.
However, explicit and strict reasoning in terms of priorities has never been elaborated in literature so far, and B-DNNs are not an exception.
Its introduction is likely to bring significant benefits in the training, as the classifier would be able to better leverage the hierarchical data structure.
In the case of an actual introduction of label-learning priorities, the HC problem would turn into an LMOP of the form as in Equation \eqref{eq:dataset_lex_hc_problem}:
\begin{customopti}|s|
    {lexmin}{\theta}{\mathcal{L}(f_1(\theta;X), Y_1),\ldots,\mathcal{L}(f_n(\theta;X), Y_n)}{}{\label{eq:dataset_lex_hc_problem}}
\end{customopti}
where $f_i$ refers to the part of the global classifier $f$ dedicated to the $i$-th hierarchy level.

Lexicographic optimization in the presence of a single, potentially multi-head, DNN is still heavily under-explored.
One of the most relevant approaches consists of projecting the side-optimization goals gradient in the null space of the batch activations \cite{kissel2020neural}.
An analogous idea was proposed in the context of reinforcement learning \cite{tercan2022solving}.
Even if effective, such techniques do not scale with the network dimensions, drastically undermining the time performance and their usability in applications.

A domain closely related to lexicographic optimization is the long-studied constrained optimization \cite{martins2021engineering}, for which a large number of (not necessarily deep learning) techniques have been proposed: from penalty methods to sequential programming, from duality-based approaches to feasible directions method \cite{gong2021bi}.
Nevertheless, none of them properly suits a general purpose LMOP as they may lack rigorous respect of the priorities, as for the penalization methods, or are not designed for non-convex problems, as Lagrange multiplier and primal-dual methods \cite{bertsekas2014constrained}.
On the contrary, a theoretically compliant and efficiently implementable approach is the bi-objective gradient descent algorithm \cite{gong2021bi}, whose main limitation is the possibility of coping with up to two levels of priority.

Hereinafter, to simplify the presentation and the readability, the notation of \eqref{eq:dataset_lex_hc_problem} will be simplified to \eqref{eq:lex_hc_problem}, where the focus is on one single data point $x$ and its set label $y = (y_1,\ldots,y_n)$.
\begin{customopti}|s|
    {lexmin}{\theta}{\mathcal{L}(f_1(\theta;x), y_1),\ldots,\mathcal{L}(f_n(\theta;x), y_n)}{}{\label{eq:lex_hc_problem}}
\end{customopti}
This choice does not undermine the generality of the study, as the use of a linear reduction function, e.g., the arithmetic mean of the data point losses would preserve all the results presented in this article.

\subsection{From lexicographic to non-standard deep learning}
To proceed toward non-standard deep learning, one needs to rewrite \eqref{eq:lex_hc_problem} as
\begin{mini}|s|
    {\theta}{\overline{\mathcal{L}}(\theta,x, y)}{}{\label{eq:ns_hc}}
\end{mini}
where $\overline{\mathcal{L}}$ is an ad-hoc non-standard function defined according to Proposition \ref{prop:ns_dhl}.
Such a result suggests an alternative approach to HC problems: a non-standard one.

\begin{proposition}
    There exists a non-standard function $\overline{\mathcal{L}}$ for which problems \eqref{eq:lex_hc_problem} and \eqref{eq:ns_hc} are equivalent.
    \begin{proof}
        Theorem \ref{theo:nsa_equivalence} guarantees that \eqref{eq:lex_hc_problem} can be equivalently rewritten in the form
        \begin{mini}|s|
            {\theta}{\sum_{i=1}^n\mathcal{L}(f_i(\theta;x), y_i)\eta^{i\um1}}{}{\label{eq:lex_ns_hc}}
        \end{mini}
        Each function $f_i$ in \eqref{eq:lex_ns_hc} is parametrized according to weights $\theta$ that can be split in three disjoint sets: $\theta_s$, i.e., parameters shared with the other $f_j$, $j\neq i$; $\theta_i$, i.e., parameters contributing to $f_i$ only; $\theta_{\neg i}$, i.e., parameters not contributing to $f_i$ at all.
        This allows the rewriting of $f_i$ as the composition of two functions, $g_i$ and $g_s$, only depending on parameters $\theta_i$ and $\theta_s$, respectively:
        \[
        f_i(\theta;x) = g_i(\theta_i, g_s(\theta_s; x))\;\;\;\forall i=1,\ldots,n.
        \]
        According to the universal approximation theorem \cite{haykin1998neural,hornik1989multilayer}, $g_s$ can be chosen sufficiently complex to guarantee that any $g_i$ is linear, i.e.,
        \begin{equation}
        f_i(\theta;x) = \langle\theta_i, g_s(\theta_s; x)\rangle\;\;\;\forall i=1,\ldots,n.\footnote{To simplify the reading, the set of parameters contributing to $f_i$ only and their organization as a vector have been represented with the same symbol $\theta_i$.}
        \label{eq:universal}
        \end{equation}
        Substituting Equation \eqref{eq:universal} into \eqref{eq:lex_ns_hc}, one gets
        \begin{mini*}|s|
            {\theta}{\sum_{i=1}^n\mathcal{L}(\langle\theta_i, g_s(\theta_s; x)\rangle, y_i)\eta^{i\um1}}{}{\label{eq:lex_ns_hc_v3}}
        \end{mini*}
        Defining $\overline{\mathcal{L}}$ as the non-standard function
        \begin{equation}
        \overline{\mathcal{L}}(\theta,x,y) \coloneqq \sum_{i=1}^n\mathcal{L}(\langle\theta_i, g_s(\theta_s; x)\rangle, y_i)\eta^{i\um1}
        \label{eq:loss_ns_dhl}
        \end{equation}
        completes the proof.
    \end{proof} 
    \label{prop:ns_dhl}
\end{proposition}

\begin{figure}[ht]
    \centering
    \includegraphics[width=.7\linewidth]{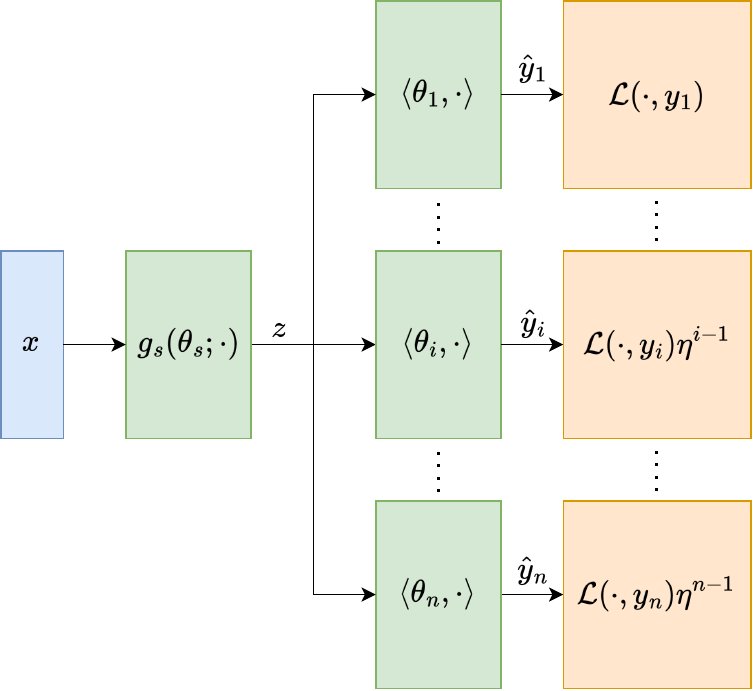}
    \caption{Representation of an LH-DNN where the per-level non-standard operations are explicitly represented. Notice that each per-level label error prediction is non-standard.}
    \label{fig:ns_dnn}
\end{figure}

The DNN induced by \eqref{eq:loss_ns_dhl} is non-standard but depends on standard weights, from here the name Hybrid in LH-DNN.
The network can be drawn as in Figure \ref{fig:ns_dnn}, where the green blocks represent the standard and non-standard sections of the LH-DNN, the orange block identifies the computation of the non-standard loss function, the light blue one is the input data.
Figure \ref{fig:ns_dnn} also allows one to evince some topology-induced properties.
The learning can be split into two phases, provided by two different partitions of the network: the left-most part, from the input to the hidden layer $z$, and the right-most part, from $z$ to the computation of the errors $\mathcal{L}(\cdot,y_i)\eta^{i\um1}$.
The first phase can be seen as a feature learning from raw input data, a process that must necessarily be in common with the classification at each level of the hierarchy.
The second phase represents the per-layer categorization and consists of multiple independent learning flaws originating from the common hidden representation $z$ of the sample $x$ to classify.

\subsection{Implementation}
The transfer principle (Theorem \ref{theo:transfer}) guarantees that an LH-DNN can be trained by backpropagation.
In particular, the gradient $\nabla_\theta \overline{\mathcal{L}}$ is non-standard and such that the level-related parameters $\theta_i$, $i=1,\ldots,n$, are adjusted independently from the performance on the other levels, while the shared parameters $\theta_s$ are adapted considering the overall categorization quality, with the losses of higher hierarchical levels having an impact infinitely larger than those of lower ones.
This phenomenon induces learning where parameters $\theta_i$ specialize for the specific level-oriented classification task, while parameters $\theta_s$ adapt to provide the most informative data inner representation, prioritizing their usefulness for coarser categorizations.
This behavior is perfectly aligned with human hierarchical learning, where feature extraction for a correct classification at the first levels is fundamental and preparatory for finer discrimination, which contributes to learning by refining an already effective process of feature retrieval.

However, the fact that the loss function gradient $\nabla_\theta \overline{\mathcal{L}}$ is non-standard raises concerns about its naive use to update the network parameters, as it is in contrast with Theorem \ref{theo:nsa_equivalence}, as it guarantees the problems equivalence if and only if the domain, i.e., the parameters are standard values.
This seems to suggest that a more sophisticated way to propagate the non-standard gradient is needed.
More formally, there is a need to extrapolate standard improvement directions from non-standard ones.

A principled approach seems to be suggested by the results in \cite{na_ipm}, where a non-standard interior point method for lexicographic problems has been proposed and implemented.
The following theorem, here proved for the first time in literature, resumes it.
\begin{theorem}
    Let $f_1,f_2\colon\mr^n\rightarrow\mr$, $f_1,f_2\in\mathcal{C}^1(\mr^n)$, and $f\colon\mr^n\rightarrow\me$ be defined as $f(x)\coloneqq f_1(x)+f_2(x)\eta$.
    Let also $\overline{x} \in\mr^n$ be a local minimum for $f_1$ but not for $f_2$.
    Then, given any arbitrary small $\lambda\in\mr^+$, $\nabla_x f|_{\overline{x}}$ identifies a standard improving direction for $f$ at $\overline{x}$ if and only if
    \begin{equation*}
        f_1\left(\overline{x}-\lambda\nabla_{x}f_2\Big|_{\overline{x}}\right) = f_1(\overline{x}).
    \end{equation*}
    \label{theo:secondary_grad_at_opt}
    \begin{proof}
        According to the transfer principle (Theorem \ref{theo:transfer}), the non-standard gradient is a linear operator, and so it holds
        \begin{equation*}
            \nabla_x f|_{\overline{x}} = \nabla_x f_1|_{\overline{x}} + \eta\nabla_x f_2|_{\overline{x}} = \eta\nabla_x f_2|_{\overline{x}},
        \end{equation*}
        since $\nabla_x f_1|_{\overline{x}} = 0$ as $\overline{x}$ is a local minimum for $f_1$ by hypothesis.
        However, by definition, $\nabla_x f|_{\overline{x}}\in\me^n$, which means that it is a non-standard improving direction for $f$ at $\overline{x}$.
        A standard direction parallel to it is $\nabla_x f_2|_{\overline{x}}$, as the two coincide up to a non-standard scaling factor.
        Thus, $\nabla_x f|_{\overline{x}}$ identifies a standard locally improving direction for $f$ at $\overline{x}$ if and only if $\nabla_x f_2|_{\overline{x}}$ does.
        To be true, this requires that
        \begin{equation}
            f(\overline{x}-\lambda\nabla_x f_2|_{\overline{x}}) \le f(\overline{x}) = f_1(\overline{x}) + f_2(\overline{x})\eta.
            \label{eq:f_grad_dec}
        \end{equation}
        By definition,
        \begin{equation*}
        \begin{split}
            f(\overline{x}-\lambda\nabla_x f_2|_{\overline{x}}) &= f_1(\overline{x}-\lambda\nabla_x f_2|_{\overline{x}}) + f_2(\overline{x}-\lambda\nabla_x f_2|_{\overline{x}})\eta <\\
            &< f_1(\overline{x}-\lambda\nabla_x f_2|_{\overline{x}}) + f_2(\overline{x})\eta,
        \end{split}
        \end{equation*}
        as $\nabla_x f_2|_{\overline{x}}$ is a decreasing direction for $f_2$ at $\overline{x}$.
        To guarantee the satisfaction of \eqref{eq:f_grad_dec}, it is necessary that
        \begin{equation*}
            f_1(\overline{x}-\lambda\nabla_x f_2|_{\overline{x}}) \le f_1(\overline{x}),
        \end{equation*}
        however, $\overline{x}$ is a local minimum for $f_1$, i.e.,
        \begin{equation*}
            f_1(\overline{x}-\lambda\nabla_x f_2|_{\overline{x}}) \ge f_1(\overline{x}),
        \end{equation*}
        forcing the condition
        \begin{equation*}
            f_1(\overline{x}-\lambda\nabla_x f_2|_{\overline{x}}) = f_1(\overline{x}),
        \end{equation*}
        which proves the thesis.
    \end{proof}
\end{theorem}

\begin{corollary}
    Let $f_1$, $f_2$, $f$, and $\overline{x}$ be defined according to Theorem \ref{theo:secondary_grad_at_opt}.
    Then $\overline{\delta}\in S^{n-1}\cup \{0\}$ identifies the standard direction of maximum improvement for $f$ at $\overline{x}$ if and only if
    \begin{equation}
        \overline{\delta} = \arg\max_{\delta\in\Delta_{f_1}}  \left\langle \delta, \nabla_{x}f_2\Big|_{\overline{x}} \right\rangle
        \label{eq:delta}
    \end{equation}
    where
    \begin{equation*}
        \Delta_{f_1}\coloneqq\left\{\delta\in S^{n-1}\cup \{0\}\,|\,f_1(\overline{x}-\lambda\delta) = f_1(\overline{x}) \right\} 
    \end{equation*}
    for any arbitrary small $\lambda\in\mr^+$ ($S^{n-1}$ refers to the $n$-dimensional unit sphere centered in 0, as usual).
    \begin{proof}
        As already mentioned in Theorem \ref{theo:secondary_grad_at_opt}, the assumption that $\overline{x}$ is a local minimum for the standard function $f_1$ implies that
        \begin{equation*}
            f(\overline{x}-\lambda\delta) < f(\overline{x}) \Longrightarrow f_1(\overline{x}-\lambda\delta) = f_1(\overline{x}),
        \end{equation*}
        i.e., the request $\delta\in\Delta_{f_1}$ is a necessary condition for $\delta$ to be a standard improving direction (up to a standard multiplicative factor) for $f$ at $\overline{x}$.

        By construction, $\Delta_{f_1}\neq\emptyset$ as $0\in\Delta_{f_1}$.
        Moreover, $\nabla_{x}f_2|_{\overline{x}}\neq0$ as $\overline{x}$ is not an extreme point for $f_2$.
        The two imply that Equation \eqref{eq:delta} is always well-defined. 
        Furthermore, $\langle \overline{\delta}, \nabla_{x}f_2|_{\overline{x}}\rangle\ge0$ as $0\in\Delta_{f_1}$, i.e., $\overline{\delta}$ is a non-increasing direction for $f_2$ at $\overline{x}$.
        This means that
        \begin{equation*}
        \begin{split}
            f(\overline{x}-\lambda\overline{\delta}) =& f_1(\overline{x}-\lambda\overline{\delta}) + f_2(\overline{x}-\lambda\overline{\delta})\eta =\\ 
            =& f_1(\overline{x}) + f_2(\overline{x}-\lambda\overline{\delta})\eta \le f_1(\overline{x}) + f_2(\overline{x})\eta = f(\overline{x}),
        \end{split}
        \end{equation*}
        i.e., $\overline{\delta}$ is a non-increasing direction for $f$.
        
        Finally, by using a first-order approximation, which is justified by the first-order update of $\overline{x}$, it holds that
        \begin{equation*}
        \begin{split}
            f(\overline{x}-\lambda\overline{\delta}) - f(\overline{x}) &= (f_2(\overline{x}-\lambda\overline{\delta}) - f_2(\overline{x}))\eta \simeq\\
            &\simeq\left(f_2(\overline{x}) - \lambda\left\langle\overline{\delta},\nabla_{x}f_2\Big|_{\overline{x}}\right\rangle - f_2(\overline{x})\right)\eta =\\
            &=- \lambda\left\langle\overline{\delta},\nabla_{x}f_2\Big|_{\overline{x}}\right\rangle\eta,
        \end{split}
        \end{equation*}
        that is, the decrease of $f$ is linearly proportional to the inner product
        \begin{equation*}
            \left\langle\overline{\delta},\nabla_{x}f_2\Big|_{\overline{x}}\right\rangle.
        \end{equation*}
        This completes the proof, as the standard direction $\overline{\delta}$ maximizes such an inner product and so the decreasing of $f$.
    \end{proof}
    \label{cor:secondary_grad_at_opt}
\end{corollary}

        

\begin{figure}[t]
    \centering
    \includegraphics[width=.8\linewidth]{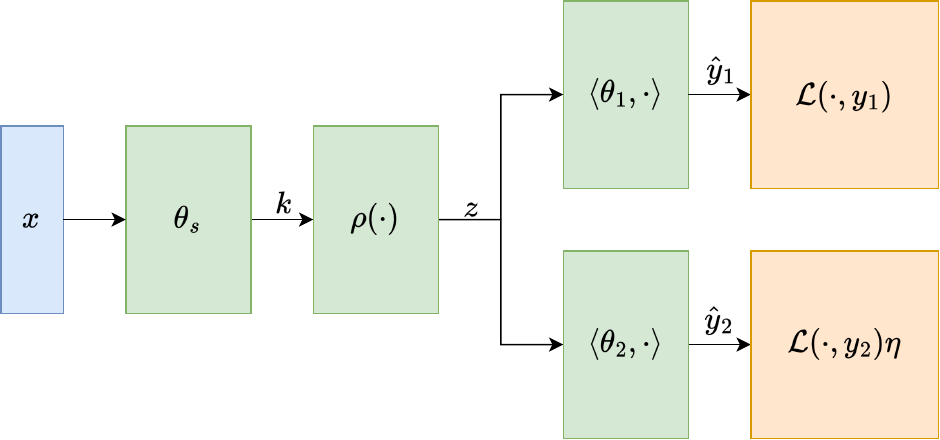}
    \caption{Representation of an LH-DNN with only one hidden layer and two objectives, i.e., two lexicographic labels for each data point. The symbol $\rho$ indicates a generic non-linearity. Such a network can be used to solve the problem in \eqref{eq:2obj_shallow}.}
    \label{fig:hdnn_simple}
\end{figure}

Such a result can be exploited to manipulate the gradient $\nabla\overline{\mathcal{L}}$ in a way that: i) $\nabla_{\theta_s}\overline{\mathcal{L}}$ is standard; ii) the contribution to the gradient of $\mathcal{L}(\cdot,y_k)$, $k>1$, is not conflicting with the losses $\mathcal{L}(\cdot,y_h)$, $h<k$.
First, the approach will be presented in the simplified version with only two objectives and where $g_s$ (remind Equation \eqref{eq:loss_ns_dhl}) consists of only one linear hidden layer.
Figure \ref{fig:hdnn_simple} graphically reports such a scenario, while Equation \eqref{eq:2obj_shallow} analytically presents it ($\rho$ indicates a generic non-linear function).\footnote{To simplify the reading, the set of shared parameters and their organization as a matrix have been represented with the same symbol $\theta_s$.}
Then, its generalization to deep multi-objective networks will be discussed.

\begin{mini}|s|
    {\theta}{\mathcal{L}(\langle\theta_1, z\rangle, y_1) + \eta \mathcal{L}(\langle\theta_2, z\rangle, y_2)}{}{\label{eq:2obj_shallow}}
    \addConstraint{z = \rho(\theta_s\,x)}{}{}
\end{mini}

\begin{theorem}
    Given the problem in \eqref{eq:2obj_shallow}, let one indicate
    \begin{equation*}
    \begin{split}
        f_1(\theta_1,\theta_s; x, y_1) = \mathcal{L}(\langle\theta_1, z\rangle, y_1),\\
        f_2(\theta_2,\theta_s; x, y_2) = \mathcal{L}(\langle\theta_2, z\rangle, y_2), 
    \end{split}
    \end{equation*}
    \begin{equation*}
        f(\theta_1,\theta_2,\theta_s; x, y_1, y_2) = f_1(\theta_1,\theta_s; x, y_1) + f_2(\theta_2,\theta_s; x, y_2)\eta.
    \end{equation*}
    Let one also indicate with $P_A$ the following (symmetric) projection operator parametrized by a full row-rank matrix (or row-vector) $A$
    \begin{equation}
        P_A \coloneqq I-A^T\left(AA^T\right)^{\um1}A = I-A^T A^{T\dagger}.
        \label{eq:prj_operator}
    \end{equation}
    Now, assume that $\mathcal{L},\rho\in C^1$ and consider a generic feasible point $(\overline{\theta}_1,\overline{\theta}_s,x,y_1)$.
    Then,
    \begin{equation}
        \overline{\delta}\coloneqq\nabla_{\theta_s}f_2\Big|_{\left( P_{\overline{\theta}_1^T\rho'_k}\overline{\theta}_2,\overline{\theta}_s,x,y_2\right)}
        \label{eq:delta_compact_def}
    \end{equation}
    is a standard improving direction for $f$ at $(\overline{\theta}_1,\overline{\theta}_2,\overline{\theta}_s,x,y_1,y_2)$, where $\rho'_k$ is the diagonal matrix obtained by computing the derivative of $\rho$ at the point $k = \langle\theta_s, x\rangle$.
    \begin{proof}
        The hypotheses satisfy those of Theorem \ref{theo:secondary_grad_at_opt}, so the thesis is true if and only if
        \begin{equation}
            f_1(\overline{\theta}_1,\overline{\theta}_s - \lambda\overline{\delta} ; x, y_1) = f_1(\overline{\theta}_1,\overline{\theta}_s; x, y_1)
            \label{eq:condition_f1}
        \end{equation}
        and $\overline{\delta}$ is such that
        \begin{equation}
            \left\langle\overline{\delta}, \nabla_{\theta_s}f_2\Big|_{\left(\overline{\theta}_2,\overline{\theta}_s,x,y_2\right)} \right\rangle \ge 0.
            \label{eq:condition_f2}
        \end{equation}
        Hereinafter, for the sake of conciseness, the labels $y_i$ are omitted.

        \textit{Verification of condition \eqref{eq:condition_f1}.} Given a generic point $u$, the first order approximation of $\rho(u-\delta)$ is 
        \begin{equation*}
            \rho(u-\delta) \simeq \rho(u)-\rho'_u\delta,
        \end{equation*}
        where $\rho'_u$ the derivative of $\rho$ at the point $u$.
        Thus, it holds true that
        \begin{equation*}
        \begin{split}
            f_1(&\overline{\theta}_1,\overline{\theta}_s - \lambda\overline{\delta} ; x) = \mathcal{L}(\langle\overline{\theta}_1, \rho(( \overline{\theta}_s- \lambda\overline{\delta}) x)\rangle) =\\
            &= \mathcal{L}(\langle\overline{\theta}_1, \rho( \overline{\theta}_s x - \lambda\overline{\delta}x)\rangle) \simeq \mathcal{L}(\langle\overline{\theta}_1, \rho(\overline{\theta}_s x) - \rho_k'\lambda\overline{\delta}x)\rangle)) =\\
            &= \mathcal{L}(\langle\overline{\theta}_1, \rho(k) - \rho_k'\lambda\overline{\delta}x)\rangle) = \mathcal{L}(\langle\overline{\theta}_1, z - \rho_k'\lambda\overline{\delta}x)\rangle) =\\
            &= \mathcal{L}(\langle\overline{\theta}_1, z - \rho_k'\lambda\overline{\delta}x)\rangle) = \mathcal{L}(\overline{\theta}_1^T z- \overline{\theta}_1^T \rho_k'\lambda\overline{\delta}x),
        \end{split}
        \end{equation*}
        and
        \begin{equation*}
            f_1(\overline{\theta}_1,\overline{\theta}_s; x) = \mathcal{L}(\overline{\theta}_1^T z).\hspace{7cm}
        \end{equation*}
        Thus, condition \eqref{eq:condition_f1} surely holds if
        \begin{equation}
            \overline{\theta}_1^T \rho_k'\overline{\delta} = 0.
            \label{eq:condition_f1_v2}
        \end{equation}
        By definition,
        \begin{equation}
            \overline{\delta} = \varepsilon \rho_k' P_{\overline{\theta}_1^T\rho_k'}\overline{\theta}_2x^T,
            \label{eq:delta_def}
        \end{equation}
        where $\varepsilon\in\mr^+$ is defined as
        \[
        \varepsilon \coloneqq \nabla_{\widehat{y}_2}\mathcal{L}\Big|_{\left(\overline{\theta}_2,\overline{\theta}_s, x\right)}.
        \]
        By plugging \eqref{eq:delta_def} into the left-hand-side of \eqref{eq:condition_f1_v2} one gets
        \begin{equation*}
            \overline{\theta}_1^T\rho_k'\varepsilon \rho_k' P_{\overline{\theta}_1^T\rho_k'}\overline{\theta}_2x^T = \varepsilon\underbrace{\overline{\theta}_1^T\rho_k'P_{\overline{\theta}_1^T\rho_k'}}_{=0}\rho_k'\overline{\theta}_2 x^T = 0,
        \end{equation*}
        guaranteeing that condition \eqref{eq:condition_f1} is satisfied thanks to the associativity of matrix multiplication and the fact that $\rho_k'$ is diagonal.

        \textit{Verification of condition \eqref{eq:condition_f2}.} By definition
        \begin{equation*}
            \left\langle\overline{\delta}, \nabla_{\theta_s}f_2\Big|_{\left(\overline{\theta}_2,\overline{\theta}_s,x\right)} \right\rangle = tr\left(\overline{\delta}^T\nabla_{\theta_s}f_2\Big|_{\left(\overline{\theta}_2,\overline{\theta}_s,x\right)}\right),
        \end{equation*}
        
        \begin{equation*}
            \nabla_{\theta_s}f_2\Big|_{\left(\overline{\theta}_2,\overline{\theta}_s,x\right)} = \varepsilon\rho_k'\overline{\theta}_2 x^T, 
        \end{equation*}
        where $tr(\cdot)$ indicates the trace of a matrix, i.e., the sum of its main diagonal elements.
        So, condition \eqref{eq:condition_f2} is equivalent to
        \begin{equation*}
            tr\left(x\overline{\theta}_2^T P_{\overline{\theta}_1^T\rho_k'}^T\rho_k'\varepsilon^2 \rho_k'\overline{\theta}_2x^T\right) \ge 0,
        \end{equation*}
        which can be rewritten as
        \begin{equation*}
            tr\left(x\overline{\theta}_2^T \rho_k'\varepsilon P_{\overline{\theta}_1^T\rho_k'} \varepsilon \rho_k'\overline{\theta}_2x^T\right) \ge 0
        \end{equation*}
        since $\rho_k'$ is diagonal, $\varepsilon$ is a scalar and $P_{\overline{\theta}_1^T\rho_k'}$ is symmetric.
        By construction, $P_{(\cdot)}$ is a projector operator, so it is positive semidefinite.
        This implies that $\langle u, P_{(\cdot)} u\rangle \ge 0$ for any real vector $u$.
        Thus,
        \begin{equation*}
        \begin{aligned}
            &tr(\overline{\theta}_2^T \rho_k'\varepsilon P_{\overline{\theta}_1^T\rho_k'} \varepsilon \rho_k'\overline{\theta}_2) \ge0 & \text{non-negative argument,}\\
            &tr(xx^T) \ge0 &\text{by construction,}\\
            &tr\left(x\overline{\theta}_2^T \rho_k'\varepsilon P_{\overline{\theta}_1^T\rho_k'} \varepsilon \rho_k'\overline{\theta}_2x^T\right) \ge 0 & \text{because of the two above.}
        \end{aligned}
        \end{equation*}
        This completes the proof.
    \end{proof}
    \label{theo:secondary_grad_at_opt_12dnn}
\end{theorem}


\begin{corollary}
    Given the hypotheses of Theorem \ref{theo:secondary_grad_at_opt_12dnn}, assume $(\overline{\theta}_1,\overline{\theta}_s,x,y_1)$ is a local minimum for $f_1$ and $\mathcal{L}$ is locally injective at $(\overline{\theta}_1,\overline{\theta}_s,x,y_1)$.
    Then, $\overline{\delta}$ defined as in \eqref{eq:delta_compact_def} is the maximum improving direction for $f$ at $(\overline{\theta}_1,\overline{\theta}_2,\overline{\theta}_s,x,y_1,y_2)$.
    \begin{proof}
        To prove the thesis, it is necessary to verify that $\overline{\delta}$ satisfies \eqref{eq:delta}.
        The fact the $\mathcal{L}$ is locally injective forces the following double implication for any generic direction $\delta$ (again, from now on all $y_i$ are omitted for brevity):
        \begin{equation*}
        \begin{split}
            f_1(\overline{\theta}_1,\overline{\theta}_s - \lambda\delta ; x) &= \mathcal{L}(\overline{\theta}_1z- \overline{\theta}_1\rho_k'\lambda\delta x) =\\
            &= \mathcal{L}(\overline{\theta}_1z) = f_1(\overline{\theta}_1,\overline{\theta}_s; x) \Longleftrightarrow \overline{\theta}_1\rho_k'\delta = 0,
        \end{split}
        \end{equation*}
        that is $\delta$ is orthogonal to $\overline{\theta}_1\rho_k'$.

        Let $\tilde\delta$ be the value defined in \eqref{eq:delta}.
        By the uniqueness of the projection, it holds that
        \begin{equation}
            \tilde\delta = \mu P_{\overline{\theta}_1^T\rho_k'}\nabla_{\theta_s}f_2\Big|_{\left(\overline{\theta}_2,\overline{\theta}_s,x\right)},
        \label{eq:delta_tilde}
        \end{equation}
        where $\mu$ is a positive scalar normalization factor.
        By manipulating Equation \eqref{eq:delta_tilde}, one gets
        \begin{equation*}
        \begin{split}
            \tilde\delta = \mu P_{\overline{\theta}_1^T\rho_k'}\nabla_{\theta_s}f_2\Big|_{\left(\overline{\theta}_2,\overline{\theta}_s,x\right)} &=
            \mu P_{\overline{\theta}_1^T\rho_k'}\varepsilon\rho_k'\overline{\theta}_2x^T =\\
            &= \mu \varepsilon\rho_k'P_{\overline{\theta}_1^T\rho_k'}\overline{\theta}_2x^T = \mu \overline{\delta},
        \end{split}
        \end{equation*}
        which proves the thesis as $\tilde\delta$ and $\overline{\delta}$ are proven to be parallel.
    \end{proof}
    \label{cor:secondary_grad_at_opt_12dnn}
\end{corollary}

\begin{figure}[ht]
    \centering
    \includegraphics[width=.9\linewidth]{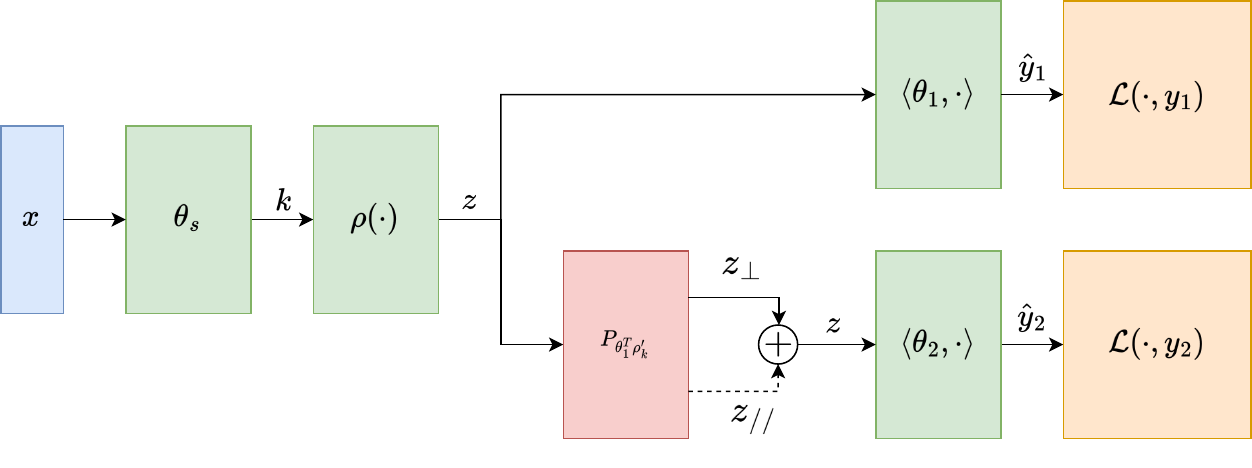}
    \caption{Representation of an LH-DNN for two-objective lexicographic learning with only one shared layer.}
    \label{fig:2obj_1layer}
\end{figure}

Theorem \ref{theo:secondary_grad_at_opt_12dnn} and Corollary \ref{cor:secondary_grad_at_opt_12dnn} are not just theoretical results.
They induce a precise network topology that guarantees the lexicographic minimization of the loss functions, i.e., the lexicographic learning of the objectives.
In Figure \ref{fig:2obj_1layer}, it is possible to appreciate such a network, as well as the major differences with the one in Figure \ref{fig:hdnn_simple}: first of all the loss functions are no more non-standard, as only standard improving directions are backpropagated through the network; such a limitation of the backpropagation is automatically managed by the projection operator of Equation \eqref{eq:prj_operator}, reported as a red block, during the forward pass.
In particular, the projection operator filters out the part of the input $z$ parallel to $\theta_1^T\rho_k'$, which is reintroduced in the flow as a constant detached from the network.
As a result, the lexicographic learning can be implemented through the usual training paradigm of a network, without any further concern on the peculiarity of the problem.

The results of Theorem \ref{theo:secondary_grad_at_opt_12dnn} and Corollary \ref{cor:secondary_grad_at_opt_12dnn} can be immediately generalized to more than two objective functions, as stated by the following proposition.
\begin{proposition}
    Consider a problem as in \eqref{eq:2obj_shallow} with $n>2$ objective functions of the form
    \begin{equation*}
        f_i(\theta_i,\theta_s,x,y_i) = \mathcal{L}(\langle\theta_i,\rho(\theta_s x)\rangle, y_i).
    \end{equation*}
    Then, the direction
    \begin{equation*}
        \nabla_{\theta_s}f_i\Big|_{(P_{A^{(i)}\rho_k'}\overline{\theta}_i,\overline{\theta}_s,x,y_i)}
    \end{equation*}
    with $i>1$ and $A^{(i)}$ defined as the matrix
    \begin{equation*}
        A^{(i)}\coloneqq 
        \begin{bmatrix}
            \overline{\theta}_1^T\rho_k'\\
            \vdots\\
            \overline{\theta}_{i\um1}^T\rho_k'
        \end{bmatrix}
    \end{equation*}
    is an improvement direction for the non-standard function $f$ defined as
    \begin{equation*}
    f(\theta_1,\ldots,\theta_n,\theta_s,x,Y) \coloneqq \sum_{i=1}^n f_i(\theta_i,\theta_s,x,y_i)\eta^{i\um1}.
    \end{equation*}
    \begin{proof}
        The proof proceeds by induction.\\
        \textit{Case $i=2$.} This case is guaranteed by Theorem \ref{theo:secondary_grad_at_opt_12dnn}.\\
        \textit{Case $i>2$.} Assume that the proposition holds for any $j=1,\ldots,i\um1$.
        Then, it also holds for index $i$ if and only if
        \begin{equation*}
            \overline{\theta}_j^T\rho_k' P_{A^{(i)}}\rho_k'\nabla_{\theta_s}f_i\Big|_{\left(\overline{\theta}_i,\overline{\theta}_s,x\right)} x^T = 0 \;\; \forall\,j=1,\ldots,i\um1,
        \end{equation*}
        which is true if and only if
        \begin{equation*}
            \overline{\theta}_j^T\rho_k' P_{A^{(i)}} = 0 \;\; \forall\,j=1,\ldots,i\um1,
        \end{equation*}
        that is $P_{A^{(i)}}(\cdot)$ is a projection operator in a space orthogonal to all the $\overline{\theta}_j^T\rho_k'$.
        Such a request holds because of the very definition of $P_{A^{(i)}}(\cdot)$, completing the proof.
    \end{proof}
    \label{prop:mobj}
\end{proposition}

\begin{figure}[ht]
    \centering
    \includegraphics[width=.9\linewidth]{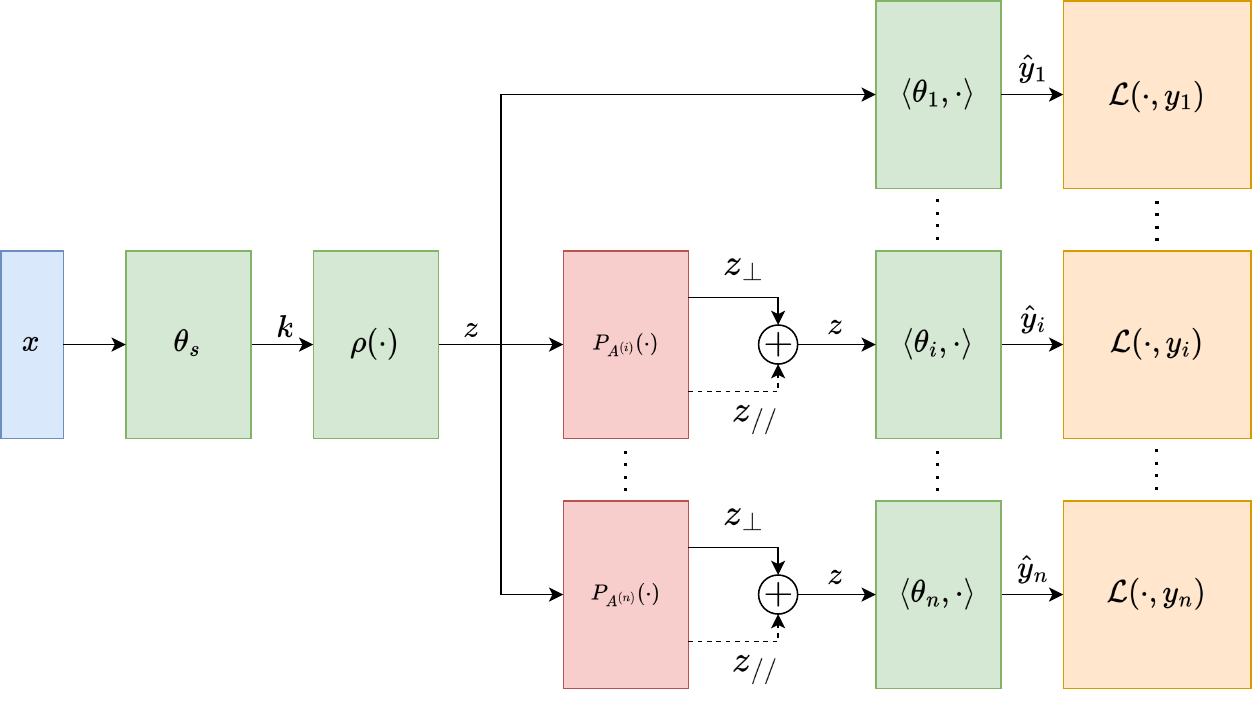}
    \caption{Representation of an LH-DNN for multi-objective lexicographic learning with only one shared layer.}
    \label{fig:mobj_1layer}
\end{figure}

Figure \ref{fig:mobj_1layer} depicts the network topology induced by Proposition \ref{prop:mobj} for such problems.
Moreover, as an implementation corollary, Proposition \ref{prop:mobj} induces a strict constraint on the dimension of $z$.
Indeed, to avoid the occurrence of trivial projections in parts of the network dedicated to finer classification, $z$ must be larger than the rows of $A^{(n)}$.
Otherwise, no gradient information would be propagated to the shared network at all.

The results of Theorem \ref{theo:secondary_grad_at_opt_12dnn} and Corollary \ref{cor:secondary_grad_at_opt_12dnn} can also be generalized for arbitrary deep networks, as stated by the following proposition.
\begin{proposition}
    Consider a problem as in \eqref{eq:2obj_shallow} with $z$ originated by the stack of $n>1$ hidden layers, that is, the composition of $n$ non-linear maps $\rho(\theta_{s_i} \cdot)$, $i=1,\ldots,n$.
    Then, indicating with $\theta_s$ the union of all $\theta_{s_i}$, the direction 
    \begin{equation*}
        \nabla_{\theta_s}f_2 \Big|_{(P_{\overline{\theta}_1^T\rho_k'}\overline{\theta}_2, \overline{\theta}_s,x,y_i)}
    \end{equation*}
    is an improving direction for $f$.
    \begin{proof}
        The proof has been moved to the supplementary material (after the references) for readability reasons.
    \end{proof}
    \label{prop:deep}
\end{proposition}

\begin{figure}[ht]
    \centering
    \includegraphics[width=.9\linewidth]{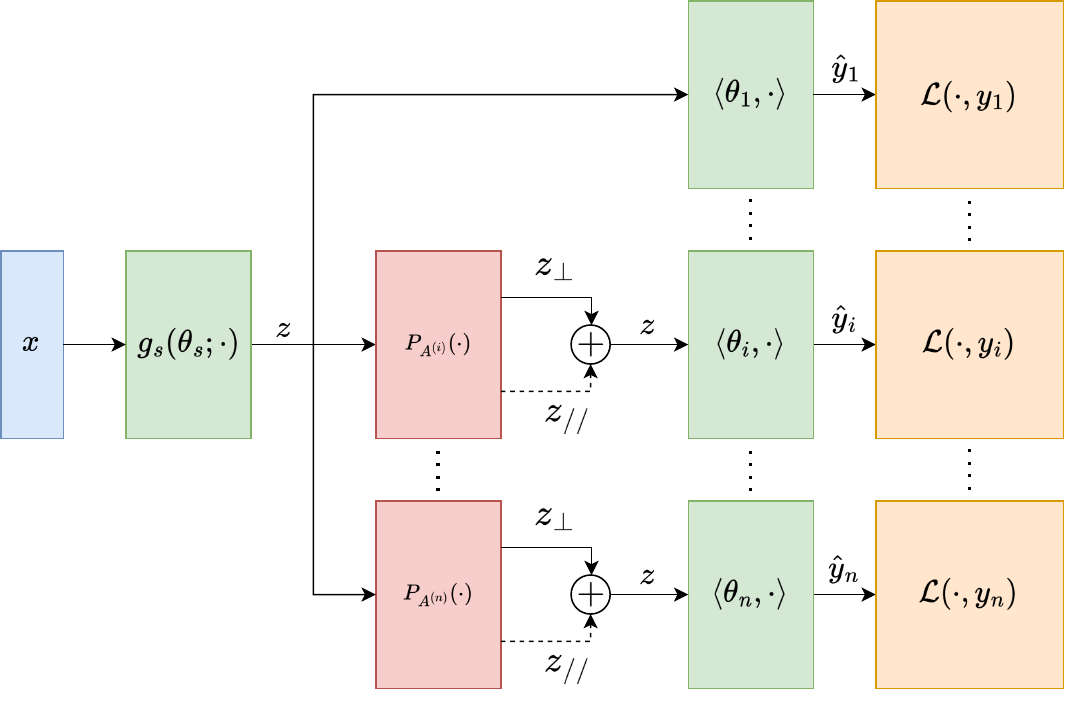}
    \caption{Compact representation of an LH-DNN.}
    \label{fig:HDNN_final_compact}
\end{figure}

Proposition \ref{prop:mobj} and Proposition \ref{prop:deep} can hold together, as stated by the next theorem.
Such a result guarantees that arbitrarily deep LH-DNNs (depicted in Figure \ref{fig:HDNN_final_compact}) can be trained through backpropagation on lexicographic multiple-objective problems.
This is possible because the two propositions are non-mutually conflicting both in the hypotheses and the theses, so they can coexist.
For this reason, the proof of Theorem \ref{theo:HDNN} is omitted, as it would replicate those of Proposition \ref{prop:mobj} and Proposition \ref{prop:deep}.
\begin{theorem}
    Arbitrarily deep multi-head neural networks as in Figure \ref{fig:HDNN_final_compact} can be trained through backpropagation for lexicographic multi-objective problems if equipped with projection operators $P_{A^{(i)}}(\cdot)$, $i=1,\ldots,n$ ($n$ is the number of objectives).
    \label{theo:HDNN}
\end{theorem}

\subsection{Interpretation and comparison with B-DNNs}
LH-DNNs possess both similarities and significant differences with respect to B-DNNs, and it is worth detailing them both for the sake of clarity and to stress the novelties in the approach presented in the current manuscript.

The most evident common aspect between the two networks is that both the topologies present branches, each dedicated to a different level of the hierarchy, departing from a shared network.
However, the branching points are different.
In a B-DNN the data flow splits at different points of the feature extraction process: the branch dedicated to a coarser classification receives a coarser representation of the data.
On the contrary, the branches separate at the same point in an LH-DNN.
This can be interpreted as a feature extraction process that improves considering the quality of the classification at different levels of the hierarchy at the same time.
The overlapping of such information is non-conflicting however, as the projection operators guarantee that the shared parameters optimization is non-detrimental for higher, coarser, and more important classifications.
This phenomenon can be seen as a feature extraction learning driven by higher-level classification tasks, with the lower one trying to improve it by contributing as much as possible.
Thus, one can expect a more efficient and effective use of the shared parameters $\theta_s$ by LH-DNNS when compared with B-DNNs.

Another analogy among the two approaches is that a loss function is computed for each level of the hierarchy separately.
Nonetheless, B-DNNs manipulate such losses through a time-varying scalarization before backpropagating them.
On the other hand, LH-DNNs preserve and do not shrink the error made for each categorization, resulting in a fairer approach.
Moreover, B-DNNs turn out to be a heavily parameter-dependent and parameter-sensitive approach due to such a scalarization.
The choice of how frequently and smoothly vary the weights is crucial for the achievement of good performance.
On the contrary, LH-DNNs constitute a parameter-less paradigm under this perspective, the reason for which it can implement more principled and robust learning.

Finally, the complexity of each branch significantly differs between B-DNNs and LH-DNNs.
In the first approach, the specialized sub-networks are assumed to be arbitrarily deep by design.
In the second one instead, it is sufficient to have just one linear layer, i.e., a linear classifier over the features extracted by the shared network.
From a theoretical perspective, this is not a limitation at all, as seen in this section.
From a computational point of view, however, there might be limitations when the correlation of the objectives is scarce and a good feature extraction for all of them might require a relatively complex shared network.
In such a case, nothing prevents further generalizing the topology, demanding to each branch the implementation of a final specialized level-wise feature extraction process by equipping them with multiple hidden layers.

\section{Experiments}
\label{sec:experiments}
The experiments involve three benchmarks: two upon which the efficacy of B-DNNs has been originally tested, namely CIFAR10 and CIFAR100 \cite{krizhevsky2009learning}, and one where B-DNNs have been subsequently used \cite{seo2019hierarchical}, namely Fashion-MNIST \cite{xiao2017}.
The experiments were realized using a machine equipped with a GraceHopper CPU (GH200), which was disposed of 72 Arm Neoverse V2 cores and an Nvidia GPU H100, with 480 GiB of memory (shared between CPU and GPU).
Coherently with the original studies, the batch size was set to 128 and the optimizer used was SGD for all the studies.
Similarly, the B-DNNs scalarizing weights, learning rates, and their switching points have been preserved as in the original study.
Thus, they are omitted for the sake of brevity.
Finally, since the B-DNNs used are mainly constituted by convolutional layers, they are typically referred to as branching convolutional NNs (B-CNNs in brief) \cite{zhu2017b}.
This convention will be maintained for the remaining of the work.

\subsection{CIFAR10 and CIFAR100}
In the original work where B-CNNs have been proposed \cite{zhu2017b}, the two benchmarks have been modified to originate a three-level hierarchical problem, whose graph has been preserved in the present study and which can be retrieved in the original manuscript.
In particular, for CIFAR10 the coarsest classification discriminates between two classes, the second one among seven, while the finest one among 10.
Similarly, the hierarchy of CIFAR100 contains eight classes in the coarsest level, 20 classes in the second level, and 100 classes in the finest one.

\begin{figure}[ht]
    \centering
    \includegraphics[width=.8\linewidth]{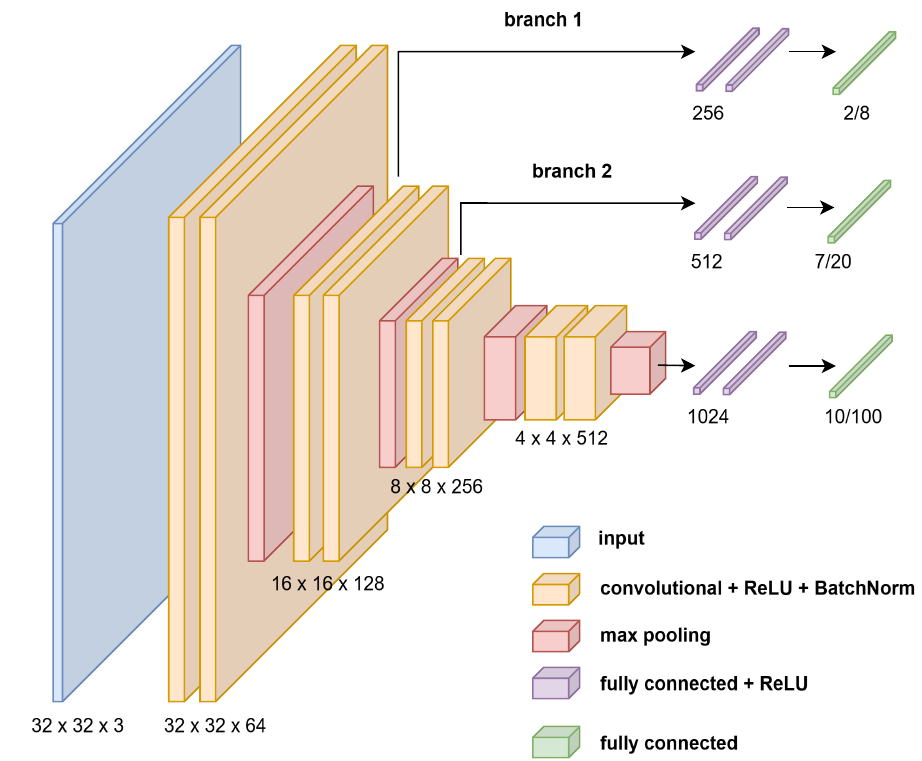}
    \caption{B-CNN used on CIFAR10 and CIFAR100.}
    \label{fig:bcnn}
\end{figure}

The B-CNN used for the study is the same as the original paper.
In particular, the network topology is reported in Figure \ref{fig:bcnn} and is heavily inspired by the work of \cite{simonyan2014very}.
The shared network consists of four pairs of $3\times3$ convolutional blocks each followed by a $2\times2$ max pooling layer.
Each branch possesses three fully connected layers with relu activation function, each output a vector whose dimension is equal to the number of classes for the corresponding layer of the hierarchy.
The network is constituted by a total of 12,382,035 trainable parameters, tuned for 60 epochs on CIFAR10 and 80 on CIFAR100.

\begin{figure}[ht]
    \centering
    \includegraphics[width=.9\linewidth]{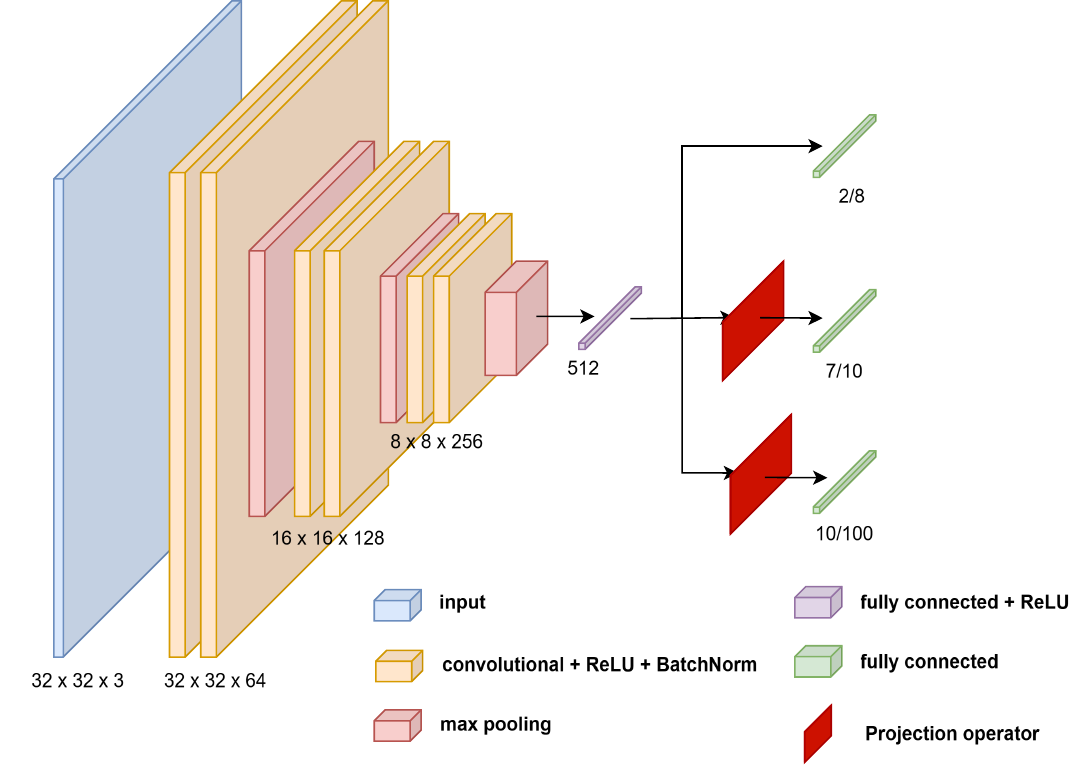}
    \caption{$s$-LH-DNN used on CIFAR10 and CIFAR100.}
    \label{fig:hcnn_c2b0}
\end{figure}

\begin{figure}[ht]
    \centering
    \includegraphics[width=.9\linewidth]{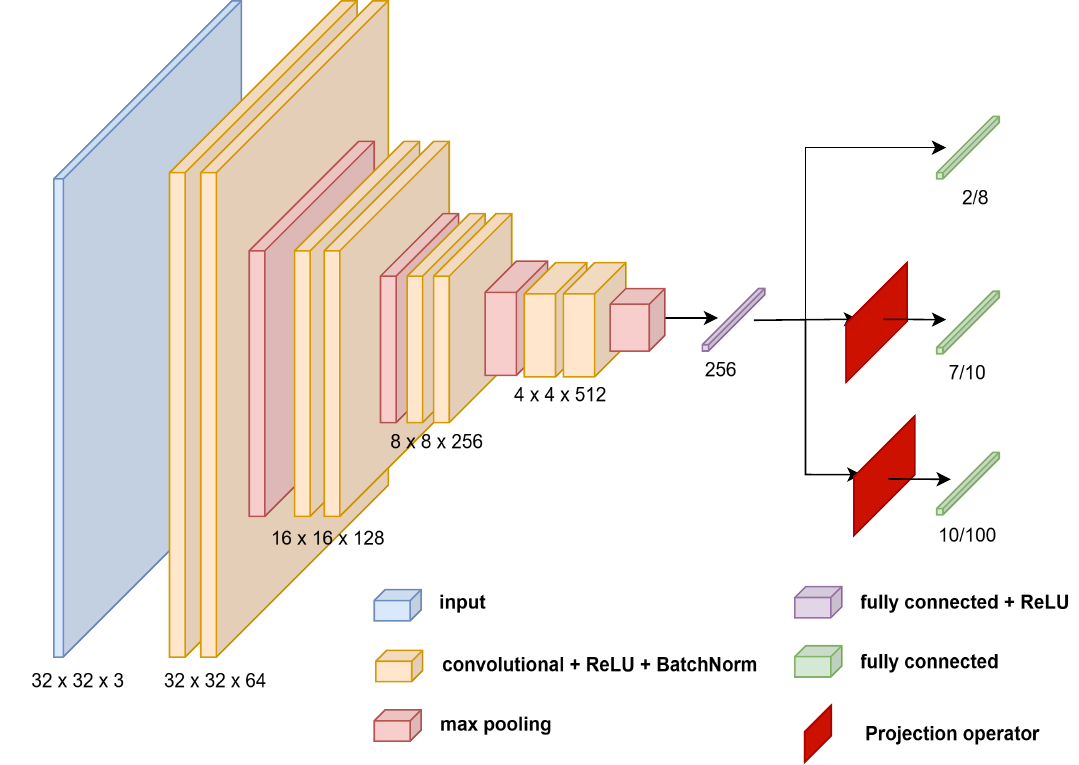}
    \caption{$\ell$-LH-DNN used on CIFAR10 and CIFAR100.}
    \label{fig:hcnn_c4b0}
\end{figure}

Concerning the LH-DNN-based approach, two networks are considered for a more comprehensive analysis: the first one ($s$-LH-DNN, where $s$ stands for ``small'') is depicted in Figure \ref{fig:hcnn_c2b0}, it consists of three pairs of $3\times3$ convolutional blocks followed by $2\times2$ max-pooling layers, and a final 256-neurons fully connected layer with relu activation function before the linear branches, for a total of only 2,200,019 parameters; the second one ($\ell$-LH-DNN, where $\ell$ stands for ``large'') is depicted in Figure \ref{fig:hcnn_c4b0}, it consists of the same shared network of the B-CNN with a final 512-neurons fully connected layer with relu activation function before the branches, for a total of  5,746,131 trainable parameters.

The choice of such topologies is meant to minimize the differences with the B-CNN so as to keep the comparison as fair as possible, rather than to boost as much as possible the performance of the LH-DNNs.
The same holds for the other hyperparameters, e.g., the LH-DDNs have been trained using the same learning rates as for the B-CNN, i.e., $3\cdot10^{\um3}$ and $5\cdot10^{\um4}$ on CIFAR10 and $10^{\um3}$ and $2\cdot10^{\um4}$ on CIFAR100.
Concerning the first benchmark, LH-DNNs have been trained for 20 epochs, and the switching of the learning rate took place at epoch 9.
For CIFAR100 instead, the two LH-DNNs have been trained for 20 and 15 epochs, respectively, and the switching of learning rate took place at epoch 11.

\begin{figure}[ht]
    \centering
    \includegraphics[width=.9\linewidth]{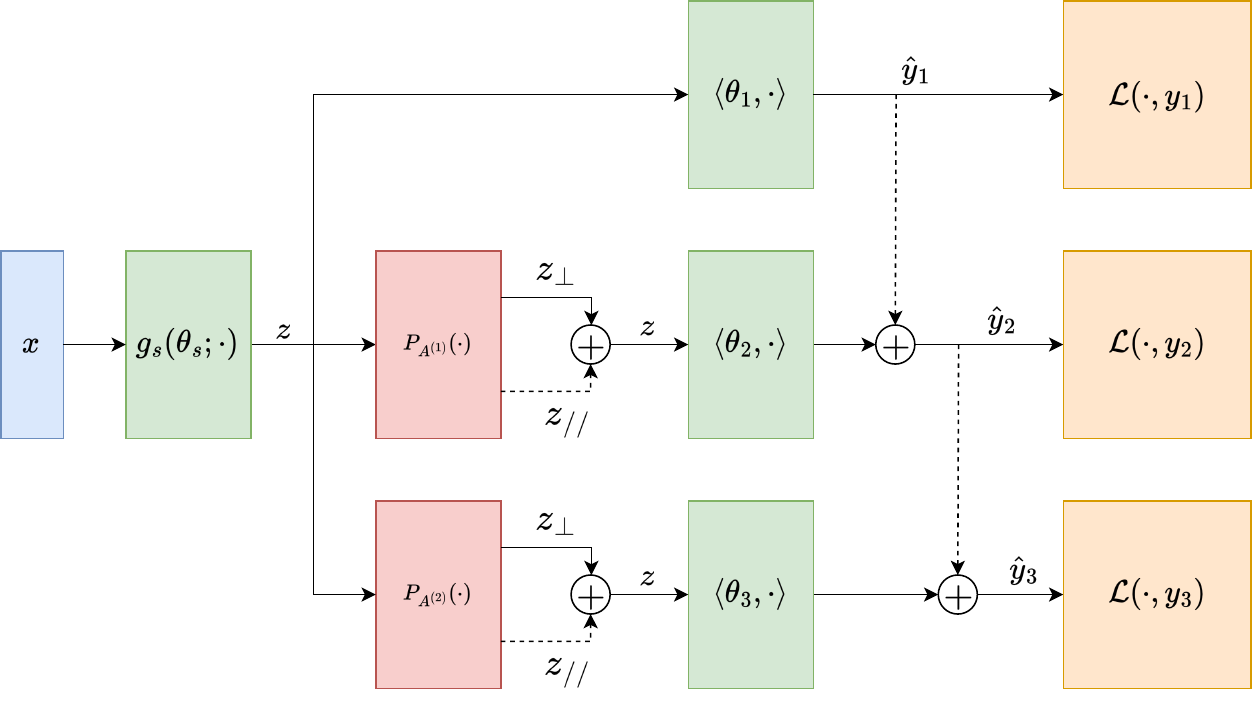}
    \caption{Schema of a generic LH-DNN for three-objective HC problem equipped with the ``advantage topology''.}
    \label{fig:hdnn_advantage}
\end{figure}

In order to ease the learning of an LH-DNN, the classification at a certain level of the hierarchy is propagated to the subsequent one as a constant.
This choice has the purpose of making the learning more robust with respect to the tree structure.
It is inspired by a notably effective reinforcement learning technique that focuses on the learning of the so-called ``advantage'' \cite{wang2016dueling}, that is a discrepancy with respect to an already acquired baseline judgment.
Figure \ref{fig:hdnn_advantage} depicts such a design choice.

\subsection{Fashion-MNIST}
Fashion-Mnist is a hierarchical-by-design benchmark with three levels of granularity: the coarsest one presents two classes, the second six, and the third ten.
The graph structure has been preserved and can be retrieved in the original manuscript.

\begin{figure}[ht]
    \centering
    \includegraphics[width=.9\linewidth]{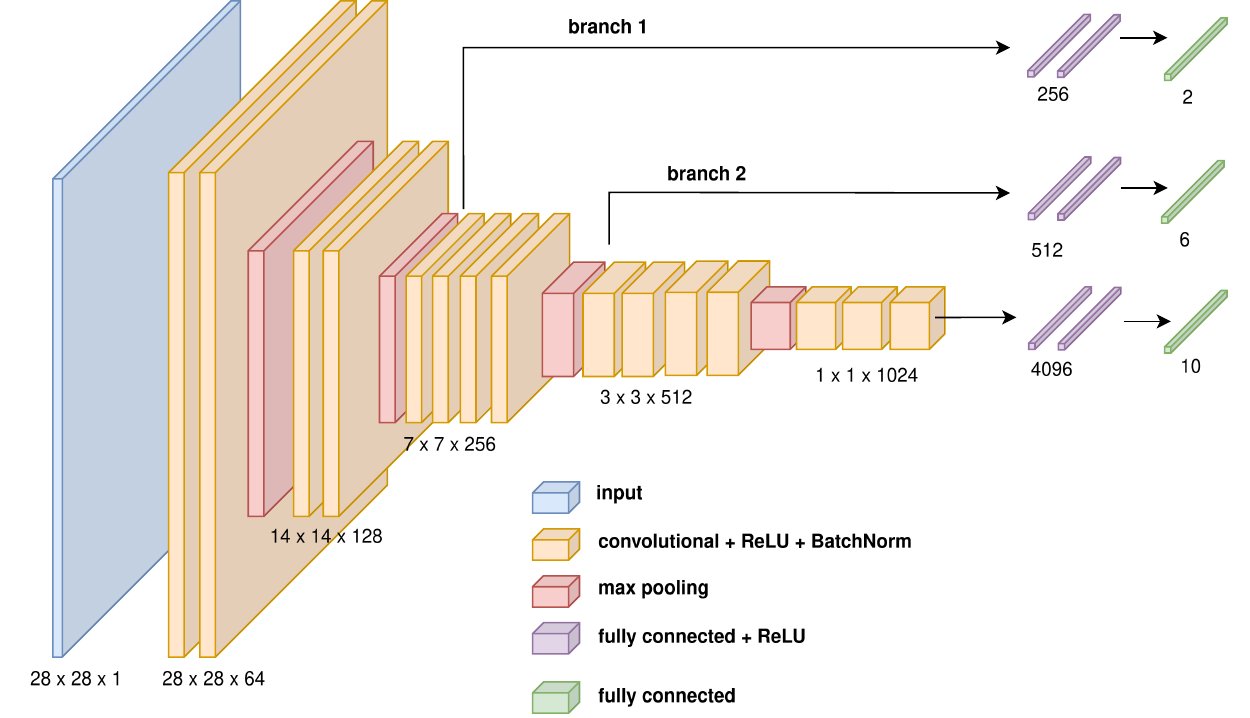}
    \caption{B-CNN used on Fashion-MNIST.}
    \label{fig:bvgg}
\end{figure}

The B-CNN used for the study, reported in Figure \ref{fig:bvgg}, is the same as in the original paper and is very similar to the previous one: it differs by two pairs of additional layers in the first and fourth $3\times3$ convolutional blocks, a fifth block of convolutional layers in the third branch, and larger fully connected layers for the third branch.
The network is constituted by a total of 38,755,282 trainable parameters.
The number of training epochs was set to 80.

\begin{figure}[ht]
    \centering
    \includegraphics[width=.9\linewidth]{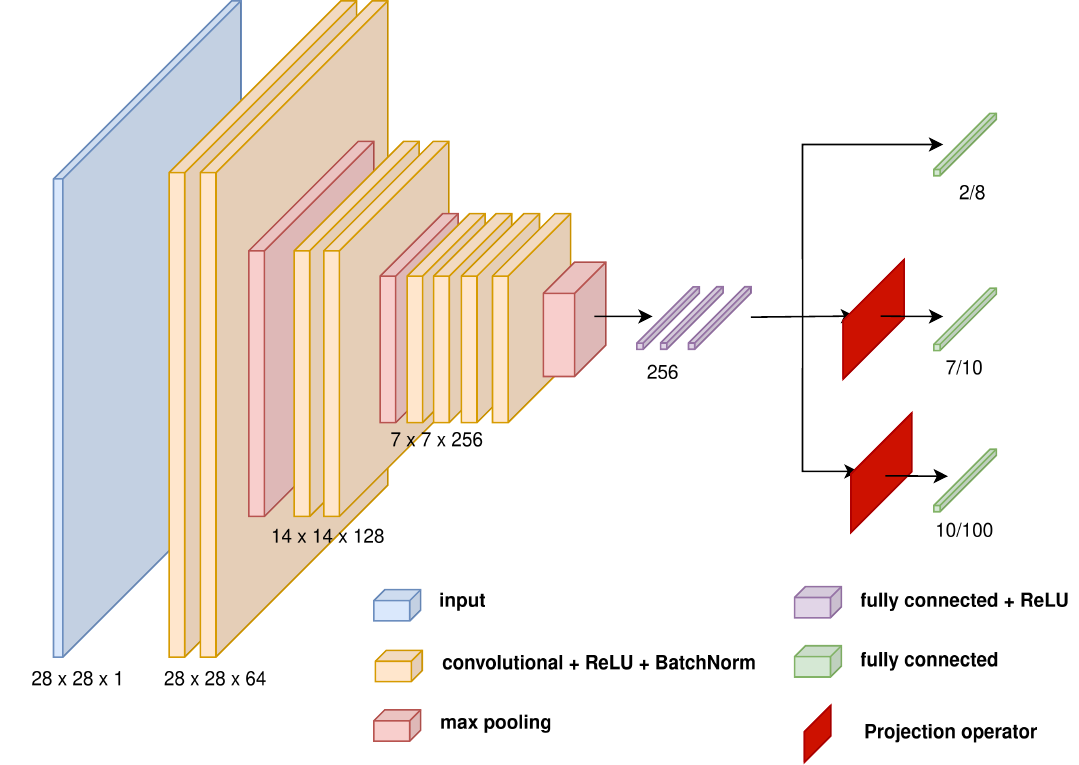}
    \caption{HDNN used on Fashion-MNIST.}
    \label{fig:lFHVGG}
\end{figure}

Concerning the LH-DNN-based approach, the one in Figure \ref{fig:lFHVGG} is considered. It consists of the very same first three blocks of the B-CNN shared part and three final 256-neurons fully connected layers with relu activation function before the linear branches, for a total of only 2,462,290 trainable parameters.

As for the other two benchmarks, the choice of such topology is meant to minimize the differences with the B-CNN so as to keep the comparison as fair as possible, rather than to boost as much as possible the performance of the LH-DNN.
The training lasted for 20 epochs using almost the same learning rates as for the B-CNN, i.e., $\cdot10^{\um3}$, $\cdot10^{\um4}$ and $5\cdot10^{\um5}$ with switching at epochs 4 and 10.
Finally, the LH-DNN has been equipped with the ``advantage topology'' as per the previous two benchmarks.

\subsection{Results on CIFAR10}
Table \ref{tab:performance_cifar10} reports the performance of the three networks, which are constituted by six metrics gathered in two macro-metrics: the accuracy and the coherency of the labeling.
The accuracy represents the quality of the classification and decomposes in an efficacy index for each level of the hierarchy.
The coherency indicates how frequently the labels assigned at each hierarchy level respect the parent-child relationship, i.e., given a data point to classify, the labeling coherency reports the probability that first and second-level labels output by a network respect the hierarchy graph; the same holds for second and third-level labeling as well as for the first and third-level one.
\begin{table}[ht]
    \centering
    \caption{Performance on CIFAR10.}
    \resizebox{\columnwidth}{!}{
    \begin{tabular}{c|c|c|c|c|c|c|}
    \cline{2-7}     & \multicolumn{3}{|c|}{\textbf{Accuracy}} & \multicolumn{3}{|c|}{\textbf{Coherency}} \\
    \cline{2-7} & $\mathbf{1}^{\text{\textbf{st}}}$ \textbf{level} & $\mathbf{2}^{\text{\textbf{nd}}}$ \textbf{level} & $\mathbf{3}^{\text{\textbf{rd}}}$ \textbf{level} & $\mathbf{1}^{\text{\textbf{st}}}$ \textbf{vs} $\mathbf{2}^{\text{\textbf{nd}}}$ & $\mathbf{2}^{\text{\textbf{nd}}}$ \textbf{vs} $\mathbf{3}^{\text{\textbf{rd}}}$ & $\mathbf{1}^{\text{\textbf{st}}}$ \textbf{vs} $\mathbf{3}^{\text{\textbf{rd}}}$ \\
    \hline
    \multicolumn{1}{|c|}{\textbf{B-CNN}} & 96.22\% & 87.51\% & 
84.12\% & 96.87\% & 93.00\% & 96.64\% \\
    \hline
    \multicolumn{1}{|c|}{\textbf{$s$-LH-DNN}} & 97.20\% & 87.67\% & 83.43\% & 98.57\% & 96.83\% & 98.45\% \\
    \hline
    \multicolumn{1}{|c|}{\textbf{$\ell$-LH-DNN}} & \cellcolor[HTML]{b7e1cd}97.36\% & \cellcolor[HTML]{b7e1cd}88.74\% & \cellcolor[HTML]{b7e1cd}84.59\% & \cellcolor[HTML]{b7e1cd}98.96\% & \cellcolor[HTML]{b7e1cd}97.84\% & \cellcolor[HTML]{b7e1cd}98.80\% \\
    \hline
    \end{tabular}}
    \label{tab:performance_cifar10}
\end{table}

From Table \ref{tab:performance_cifar10}, it turns out that an LH-DNN-based approach is effective for prioritized learning tasks.
Indeed, $\ell$-LH-DNN consistently outperforms the B-CNN on both the accuracy and coherency perspective.
This fact is remarkable considering that $\ell$-LH-DNN uses half of the B-CNN parameters and converges in less than one-third of the training epochs. 
Even more importantly, the $s$-LH-DNN achieves performances that are better than or comparable to B-CNN ones with only one-sixth of the parameters.
This fact suggests that the topology induced by NSA is able to better exploit the network parameters, efficiently and effectively adapting them for all the objectives without deteriorating the performance of the more important ones.

\begin{figure}
    \centering
    \begin{subfigure}[b]{.79\linewidth}
    \centering
    \includegraphics[width=\linewidth]{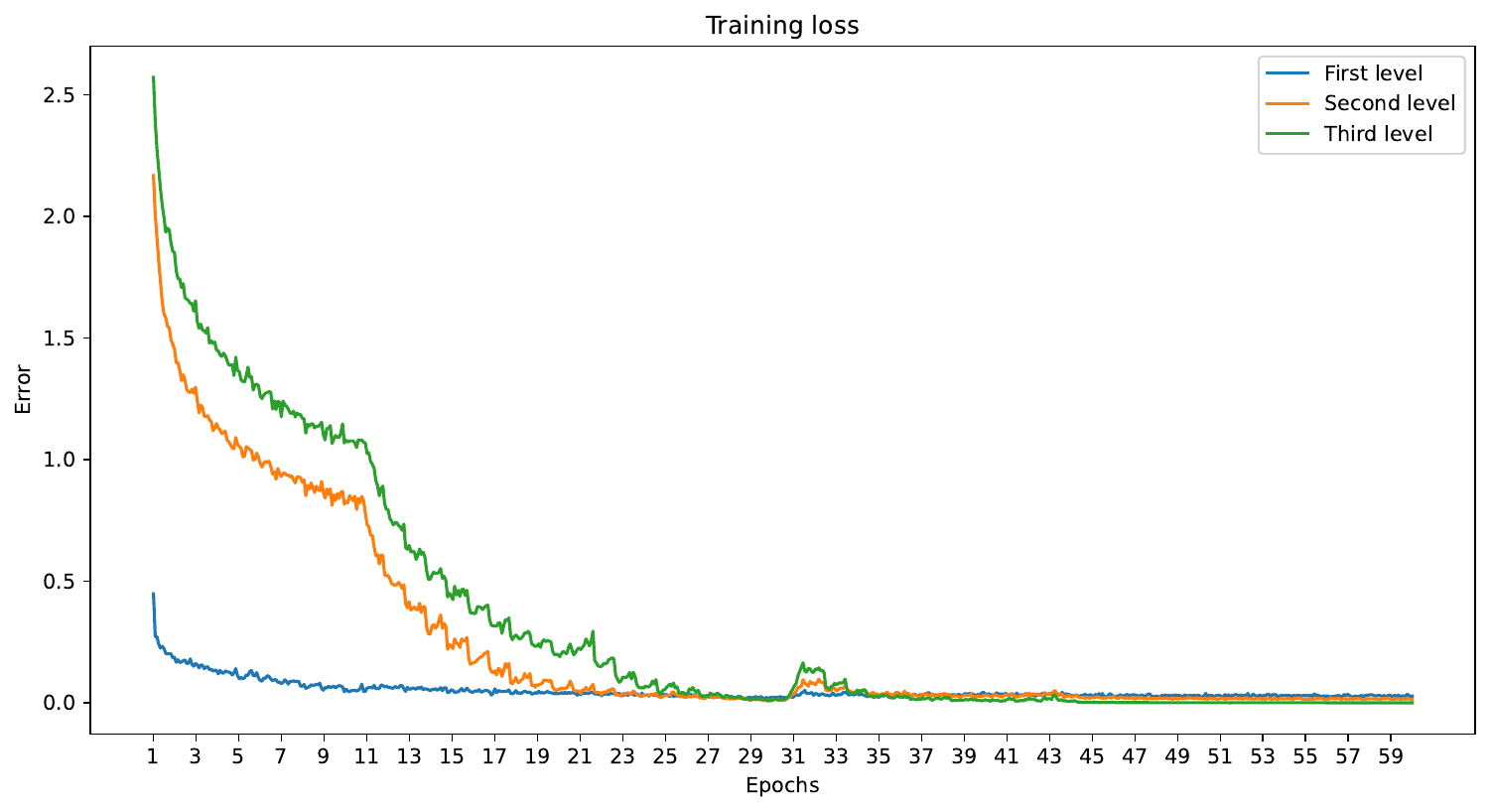}
    \caption{B-CNN train loss.}
    \label{fig:C10_B_train}
    \end{subfigure}
\\
    \begin{subfigure}[b]{.79\linewidth}
    \centering
    \includegraphics[width=\linewidth]{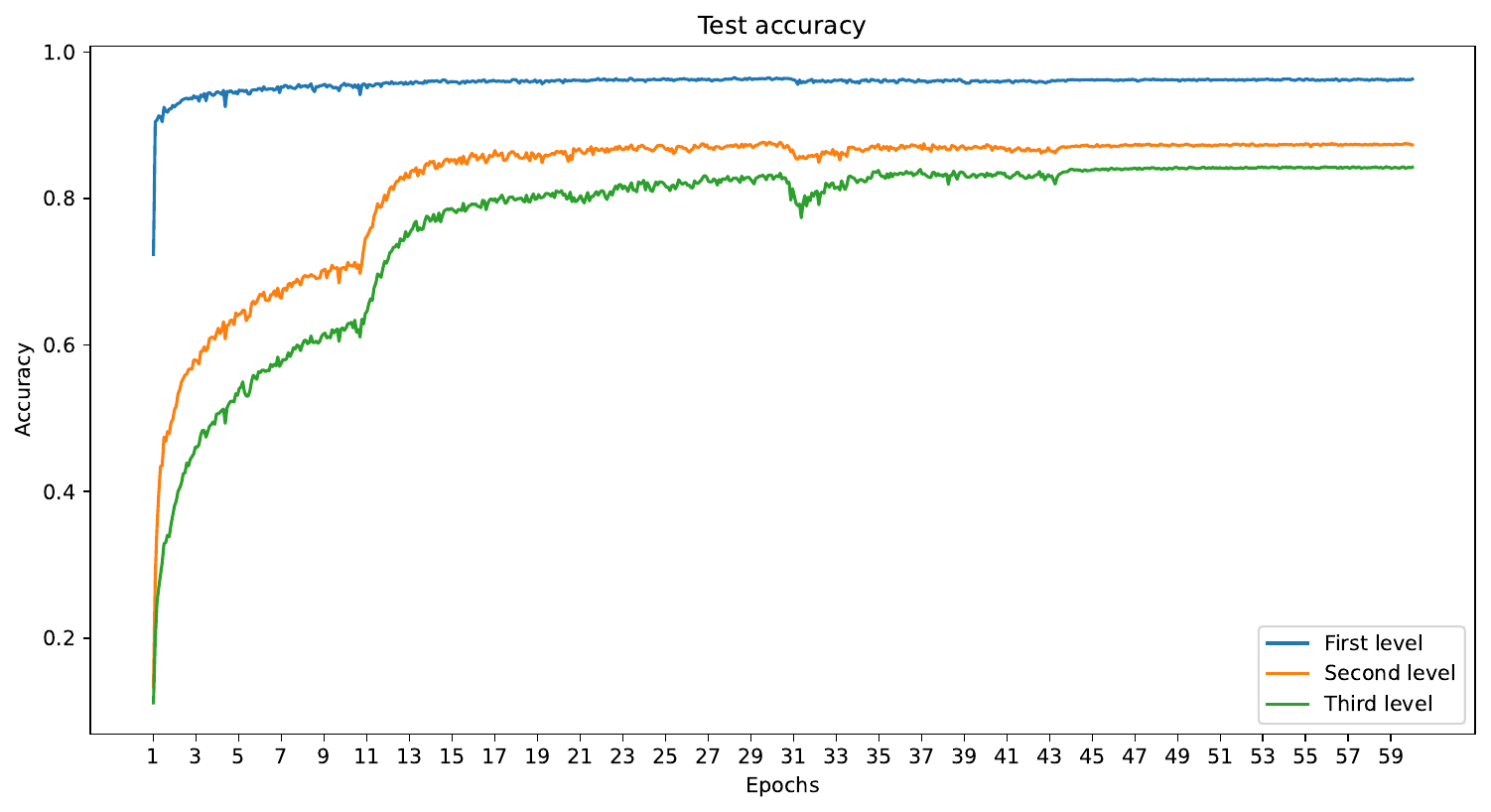}
    \caption{B-CNN test accuracy.}
    \label{fig:C10_B_test}
    \end{subfigure}
    \caption{B-CNN on CIFAR 10.}
    \label{fig:C10_B}
\end{figure}

\begin{figure}
    \centering
    \begin{subfigure}[b]{.79\linewidth}
    \centering
    \includegraphics[width=\linewidth]{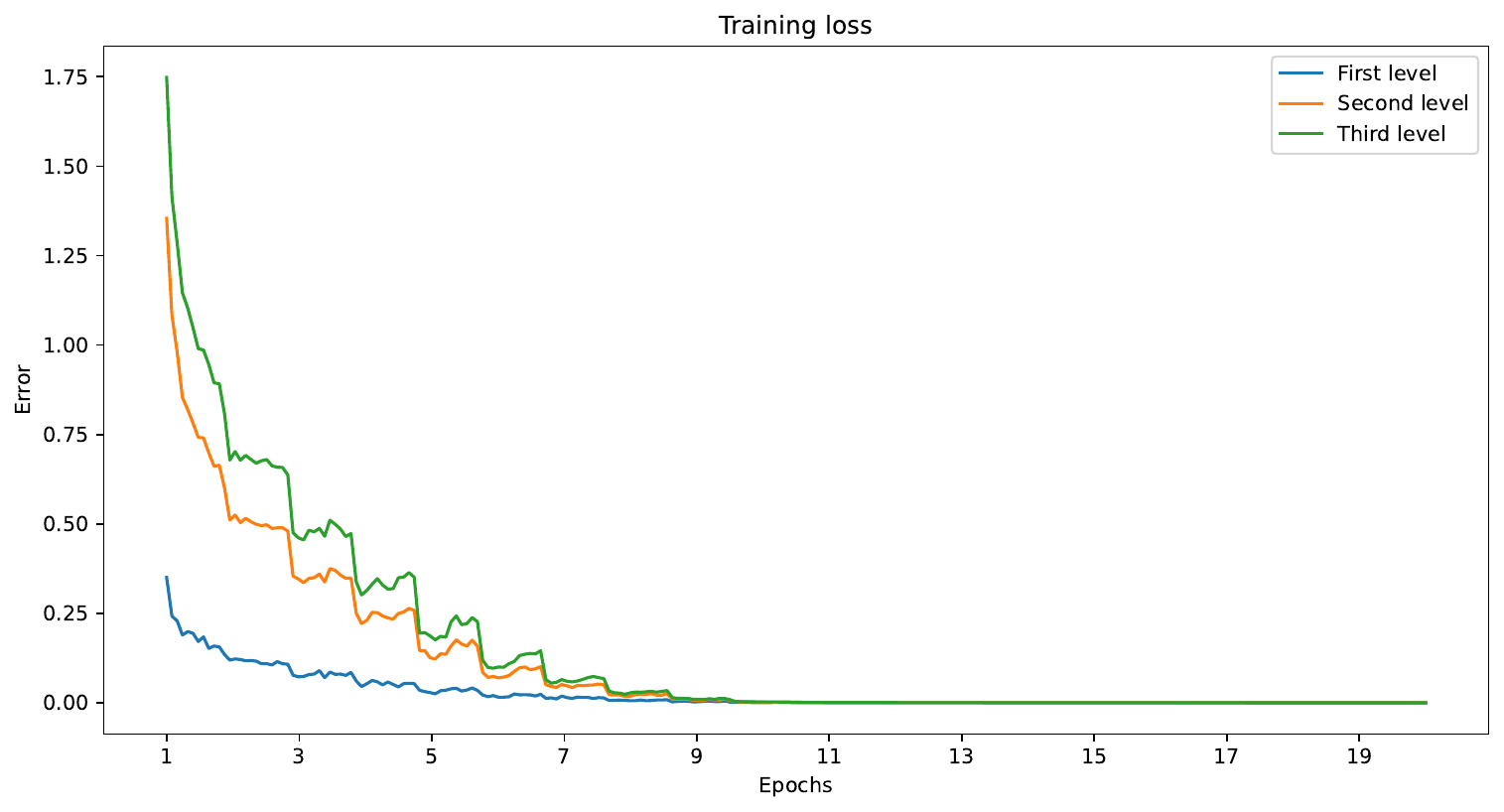}
    \caption{$s$-LH-DNN train loss.}
    \label{fig:C10_Hs_train}
    \end{subfigure}
\\
    \begin{subfigure}[b]{.79\linewidth}
    \centering
    \includegraphics[width=\linewidth]{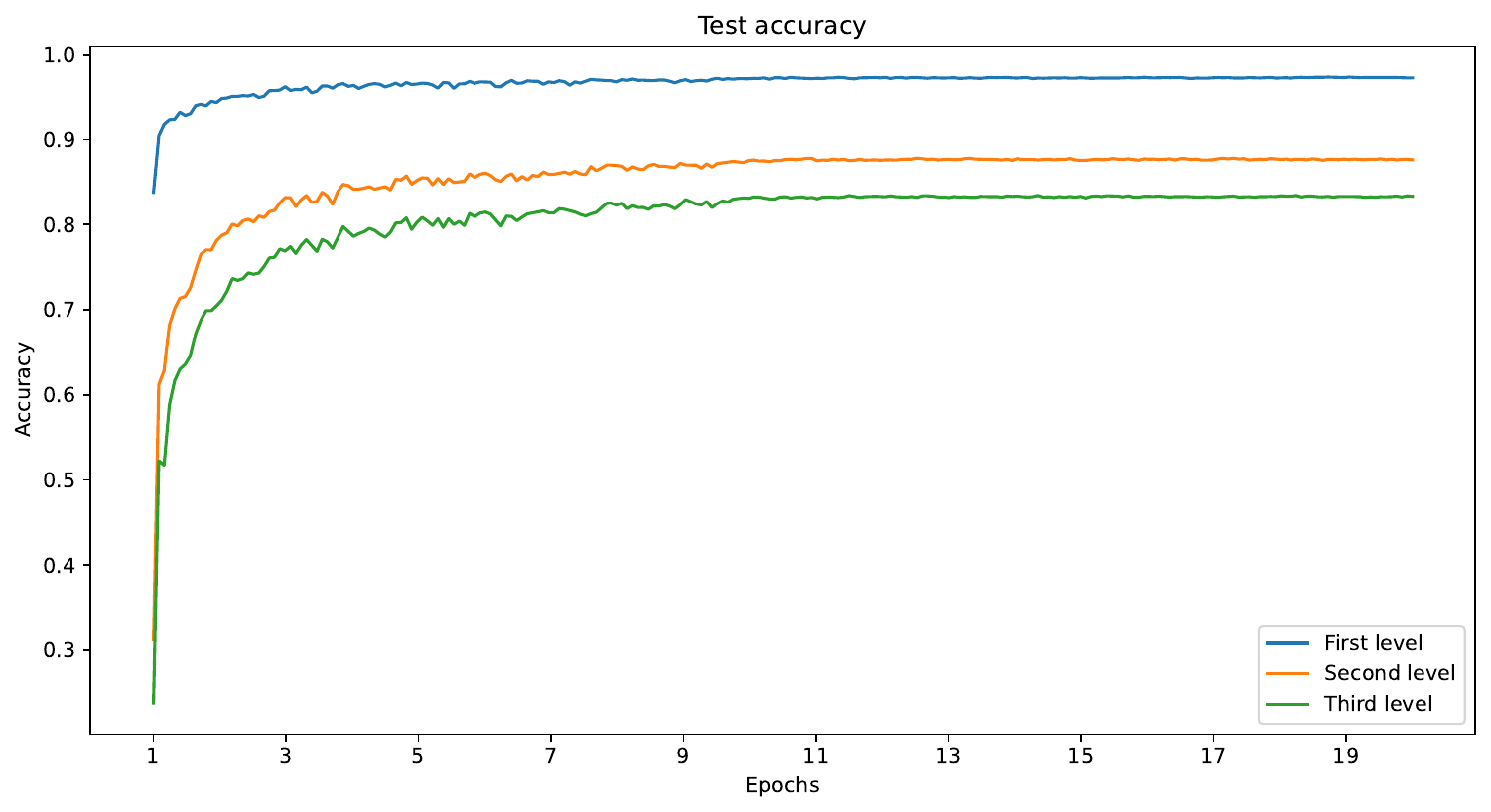}
    \caption{$s$-LH-DNN test accuracy.}
    \label{fig:C10_Hs_test}
    \end{subfigure}
    \caption{$s$-LH-DNN on CIFAR 10.}
    \label{fig:C10_Hs}
\end{figure}

\begin{figure}
    \centering
    \begin{subfigure}[b]{.79\linewidth}
    \centering
    \includegraphics[width=\linewidth]{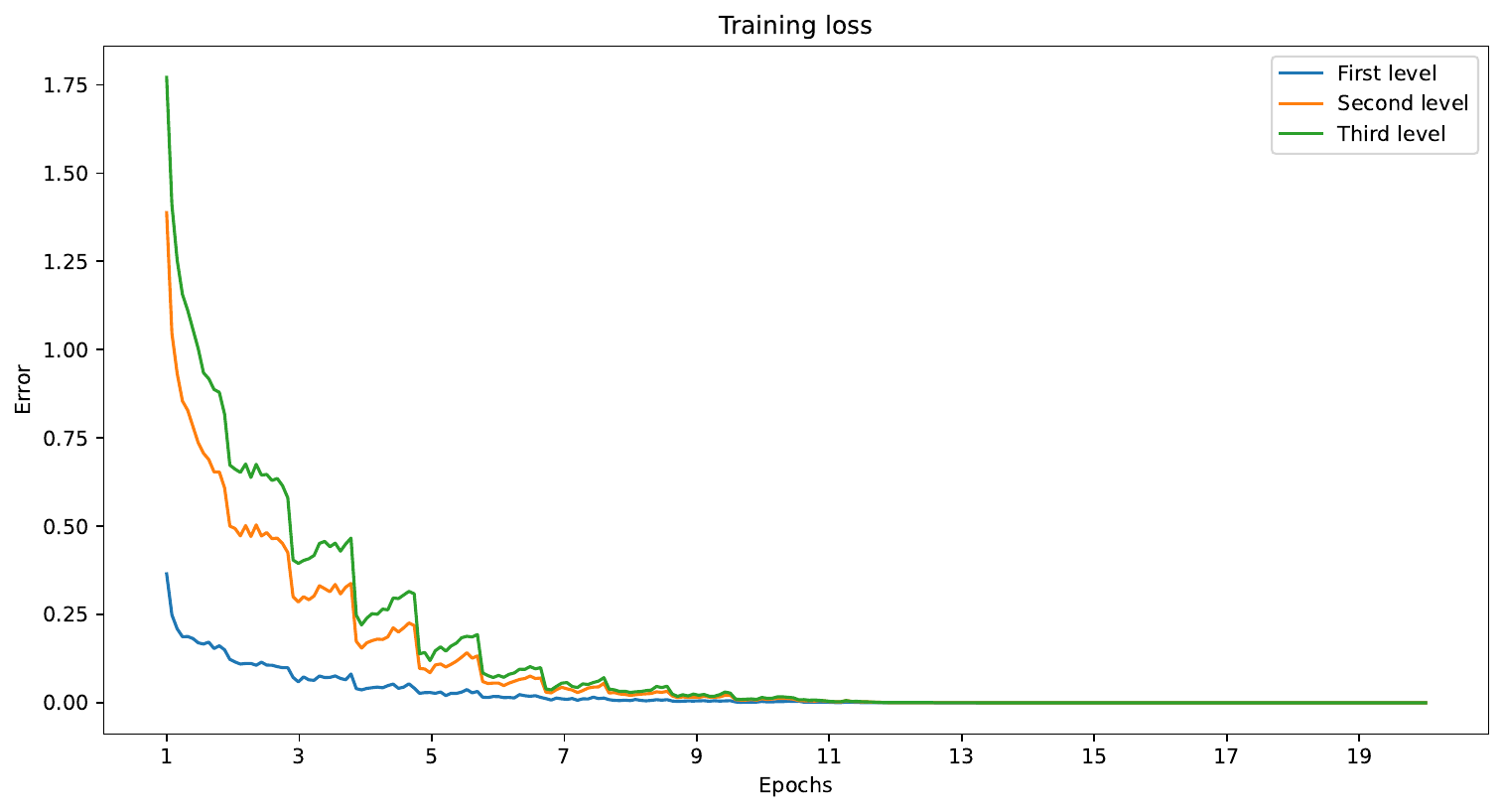}
    \caption{$\ell$-LH-DNN train loss.}
    \label{fig:C10_Hl_train}
    \end{subfigure}
\\
    \begin{subfigure}[b]{.79\linewidth}
    \centering
    \includegraphics[width=\linewidth]{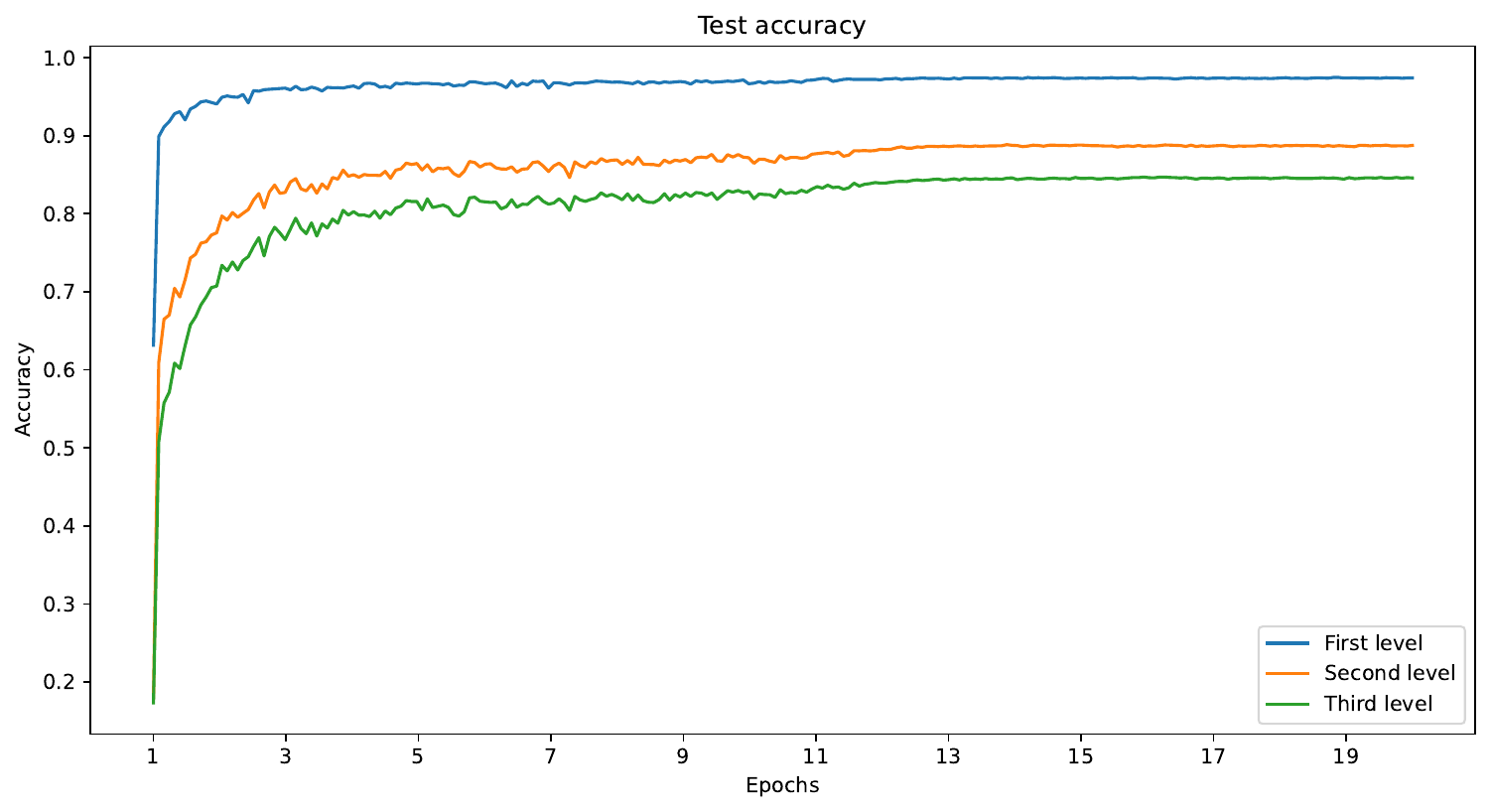}
    \caption{$\ell$-LH-DNN test accuracy.}
    \label{fig:C10_Hl_test}
    \end{subfigure}
    \caption{$\ell$-LH-DNN on CIFAR 10.}
    \label{fig:C10_Hl}
\end{figure}

Further considerations can be made by looking at Figures \ref{fig:C10_B}-\ref{fig:C10_Hl}, which report the learning error and test accuracy during the training.
The learning curve in Figure \ref{fig:C10_B} presents three degradation points: at epochs 10, 20, and 30, corresponding to the changes of the loss function scalarizing weights.
The first is minimal and manifests with increasing the first objective loss in favor of improving the second and third objectives.
This empirically testifies that a time-varying scalarization cannot a priori guarantee the respect of the hard priority among the objectives.
The second degradation takes place when the focus significantly shifts towards the third objective, which seems to be the only one affected by it.
This fact is peculiar, as the weight shift is supposed to favor it rather than penalize it, even locally.
The phenomenon repeats at epoch 30 and is even more impactful, with a performance degradation on all three objectives and a significant drop in accuracy.
Especially for the third goal, the performance downfall requires a notable amount of epochs to recover, about 15.

On the other hand, such negative events never occur during LH-DNNs training: such networks implement stabler learning, where the performance on each objective is concurrently improved always respecting the priority among them.
Evidence of this is the never-decreasing accuracy for all the objectives and the absence of loss increments of a higher priority task in the face of a loss reduction for a lower priority one (see Figure \ref{fig:C10_Hs} and \ref{fig:C10_Hl}).

\subsection{Results on CIFAR100}
Table \ref{tab:performance_cifar100} reports the performance of the three networks using the same metrics as per the study on CIFAR10.
\begin{table}[ht]
    \centering
    \caption{Performance on CIFAR100.}
    \resizebox{\columnwidth}{!}{
    \begin{tabular}{c|c|c|c|c|c|c|}
    \cline{2-7}     & \multicolumn{3}{|c|}{\textbf{Accuracy}} & \multicolumn{3}{|c|}{\textbf{Coherency}} \\
    \cline{2-7} & $\mathbf{1}^{\text{\textbf{st}}}$ \textbf{level} & $\mathbf{2}^{\text{\textbf{nd}}}$ \textbf{level} & $\mathbf{3}^{\text{\textbf{rd}}}$ \textbf{level} & $\mathbf{1}^{\text{\textbf{st}}}$ \textbf{vs} $\mathbf{2}^{\text{\textbf{nd}}}$ & $\mathbf{2}^{\text{\textbf{nd}}}$ \textbf{vs} $\mathbf{3}^{\text{\textbf{rd}}}$ & $\mathbf{1}^{\text{\textbf{st}}}$ \textbf{vs} $\mathbf{3}^{\text{\textbf{rd}}}$ \\
    \hline
    \multicolumn{1}{|c|}{\textbf{B-CNN}} & 73.42\% & 62.81\% & \cellcolor[HTML]{b7e1cd}52.51\% & 78.22\% & 70.37\% & 75.04\% \\
    \hline
    \multicolumn{1}{|c|}{\textbf{$s$-LH-DNN}} & 72.21\% & 60.76\% & 50.64\% & 90.56\% & 85.76\% & 87.43\% \\
    \hline
    \multicolumn{1}{|c|}{\textbf{$\ell$-LH-DNN}} & \cellcolor[HTML]{b7e1cd}75.07\% & \cellcolor[HTML]{b7e1cd}64.00\% & 52.47\% & \cellcolor[HTML]{b7e1cd}91.33\% & \cellcolor[HTML]{b7e1cd}87.62\% & \cellcolor[HTML]{b7e1cd}88.69\% \\
    \hline
    \end{tabular}}
    \label{tab:performance_cifar100}
\end{table}
Again, $\ell$-LH-DNN consistently outperforms the B-CNN but on the finest classification, where the performance is almost identical.
The results on the coherency are particularly remarkable, as they undoubtedly state that the B-CNN learning is not properly aware of the hierarchical structure of the labels and does not adequately leverage it during the learning.

\begin{figure}
    \centering
    \begin{subfigure}[b]{.79\linewidth}
    \centering
    \includegraphics[width=\linewidth]{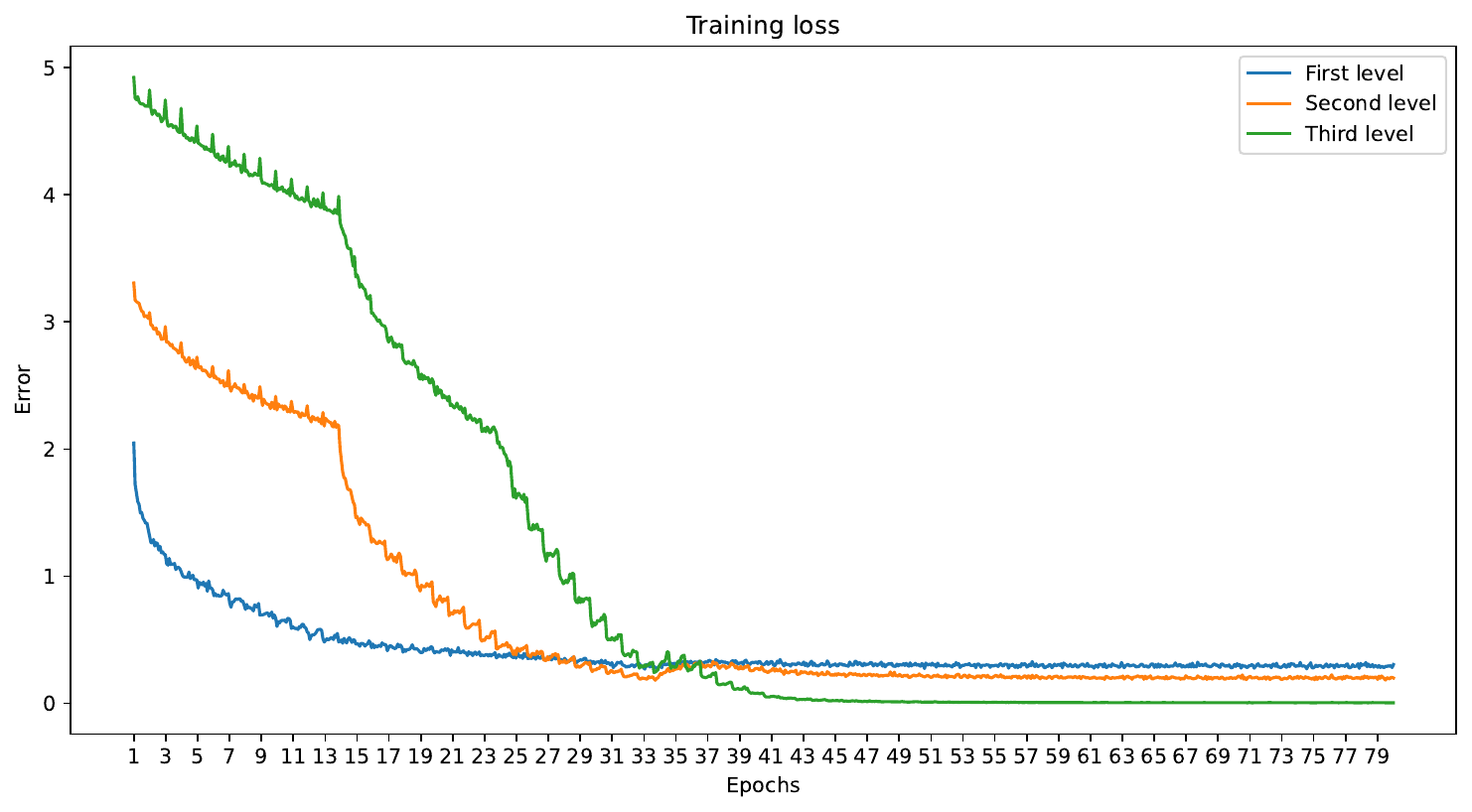}
    \caption{B-CNN train loss.}
    \label{fig:C100_B_train}
    \end{subfigure}
\\
    \begin{subfigure}[b]{.79\linewidth}
    \centering
    \includegraphics[width=\linewidth]{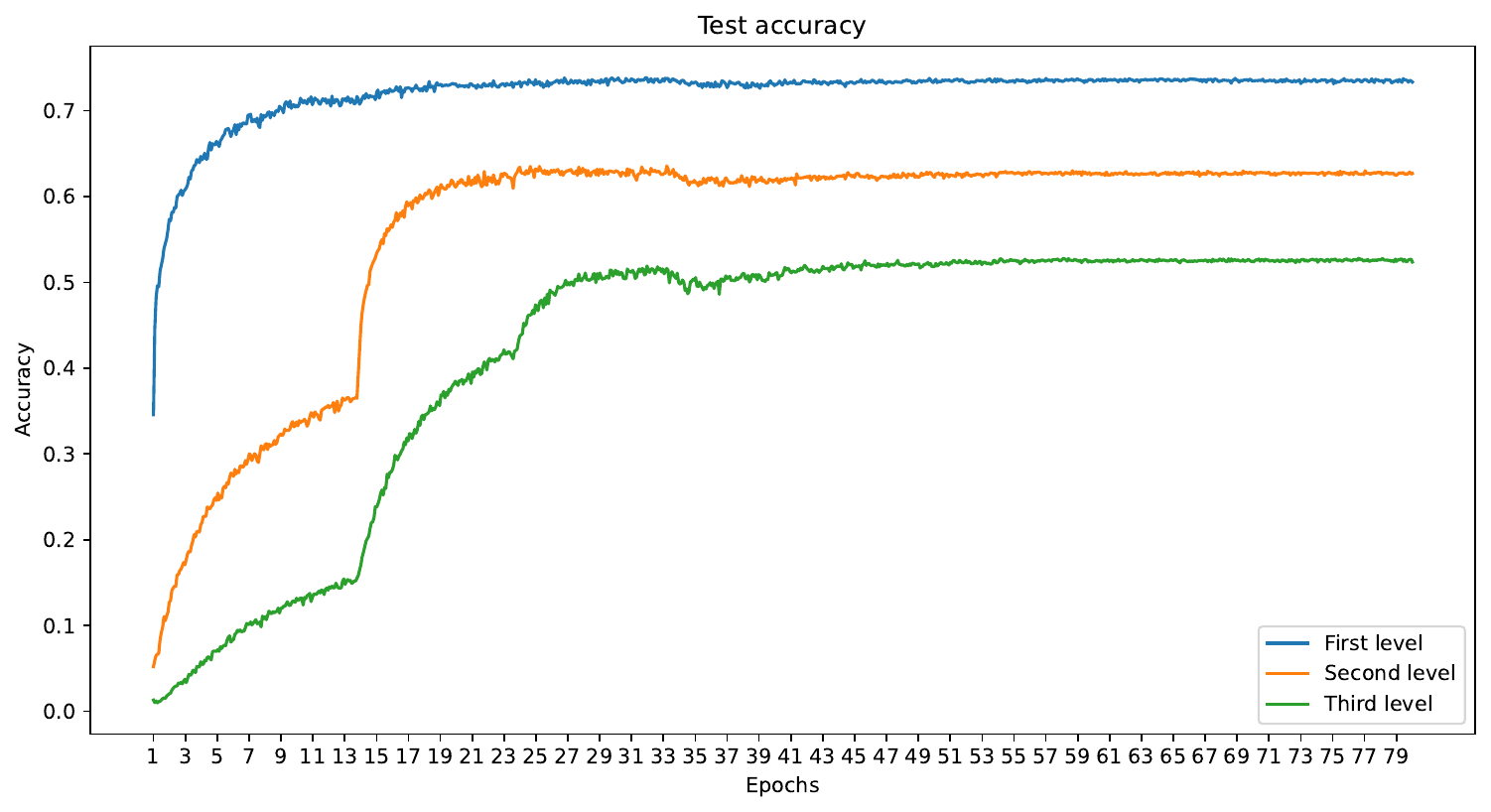}
    \caption{B-CNN test accuracy.}
    \label{fig:C100_B_test}
    \end{subfigure}
    \caption{B-CNN on CIFAR 100.}
    \label{fig:C100_B}
\end{figure}

\begin{figure}
    \centering
    \begin{subfigure}[b]{.79\linewidth}
    \centering
    \includegraphics[width=\linewidth]{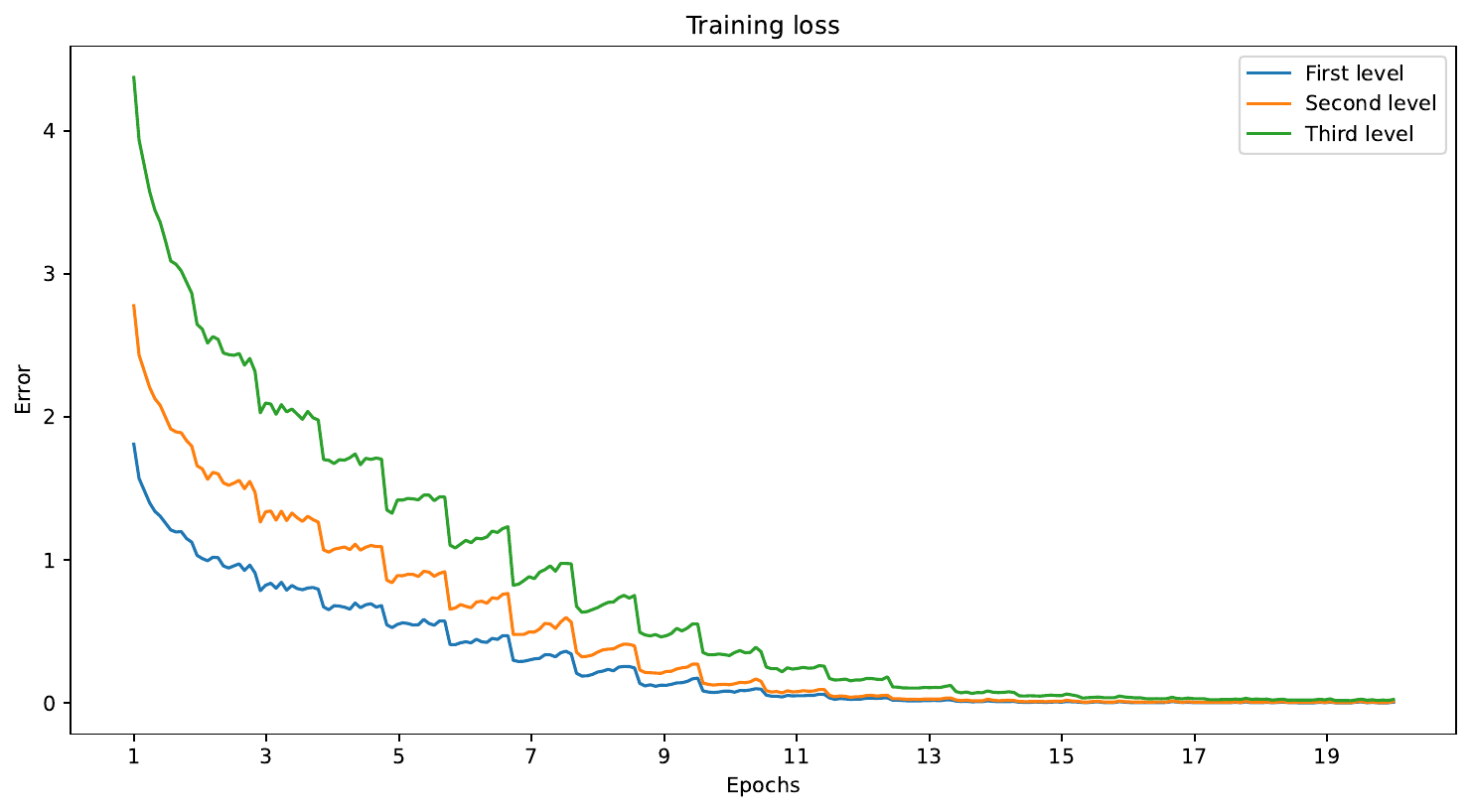}
    \caption{$s$-LH-DNN train loss.}
    \label{fig:C100_Hs_train}
    \end{subfigure}
\\
    \begin{subfigure}[b]{.79\linewidth}
    \centering
    \includegraphics[width=\linewidth]{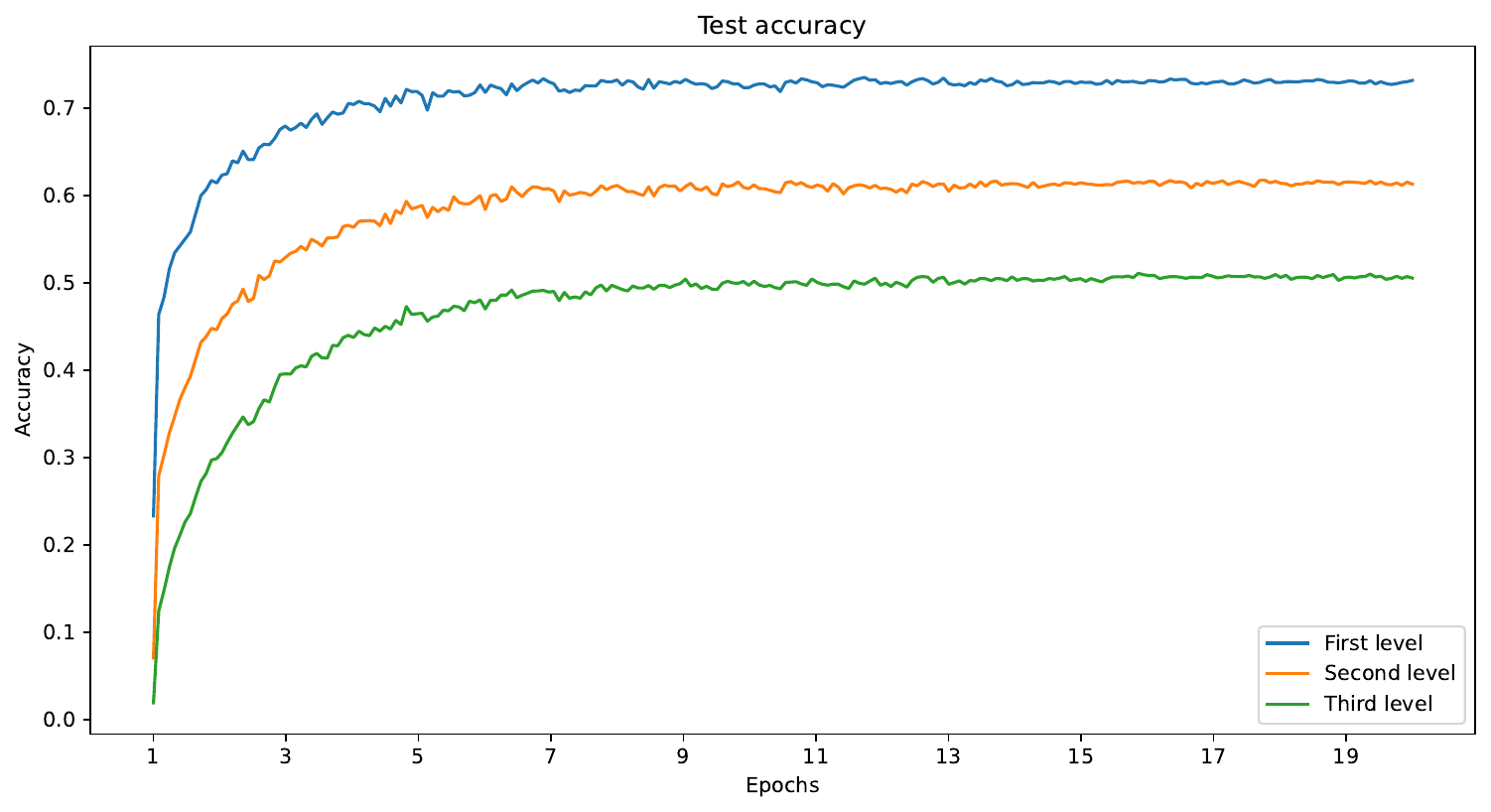}
    \caption{$s$-LH-DNN test accuracy.}
    \label{fig:C100_Hs_test}
    \end{subfigure}
    \caption{$s$-LH-DNN on CIFAR 100.}
    \label{fig:C100_Hs}
\end{figure}

\begin{figure}
    \centering
    \begin{subfigure}[b]{.79\linewidth}
    \centering
    \includegraphics[width=\linewidth]{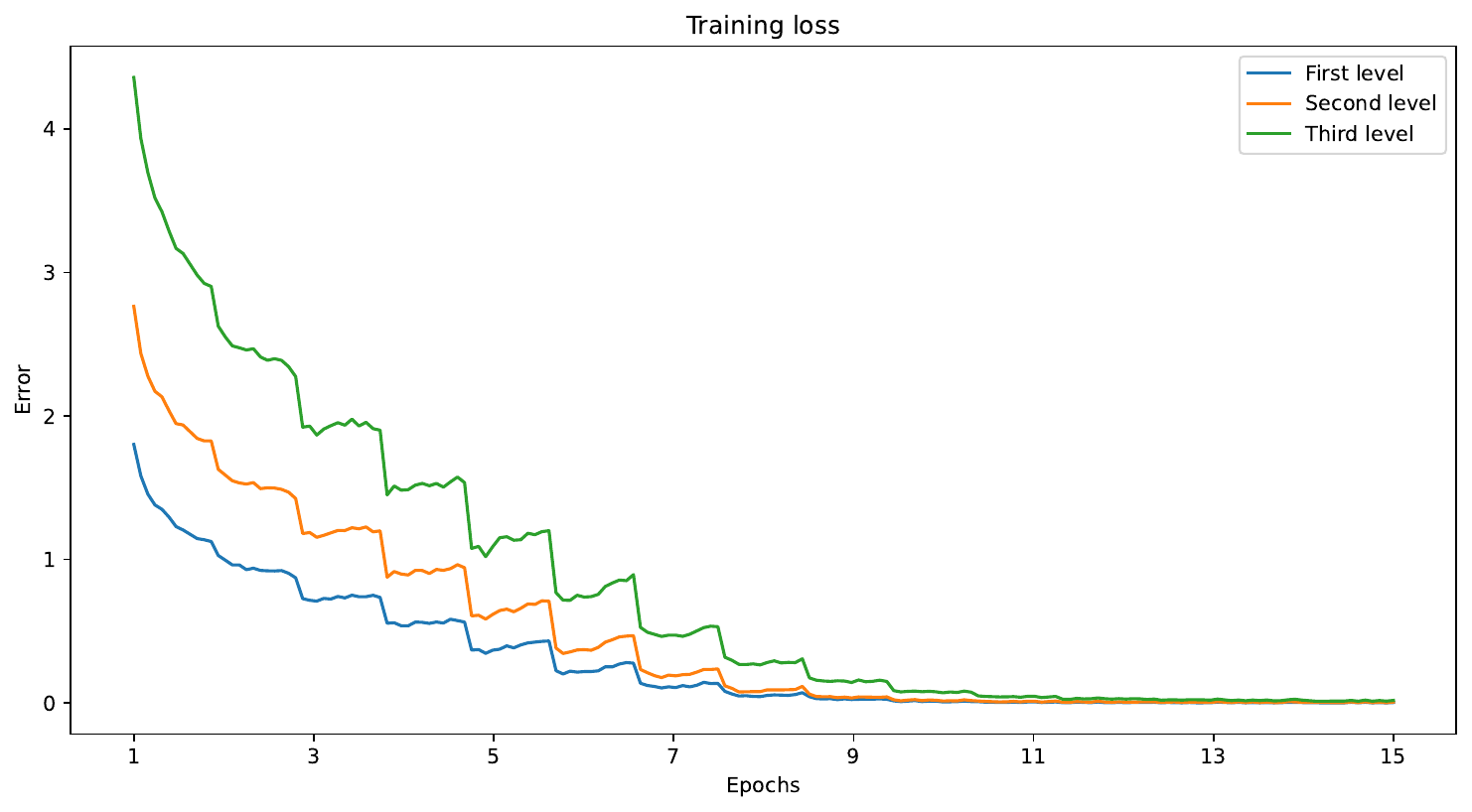}
    \caption{$\ell$-LH-DNN train loss.}
    \label{fig:C100_Hl_train}
    \end{subfigure}
\\
    \begin{subfigure}[b]{.79\linewidth}
    \centering
    \includegraphics[width=\linewidth]{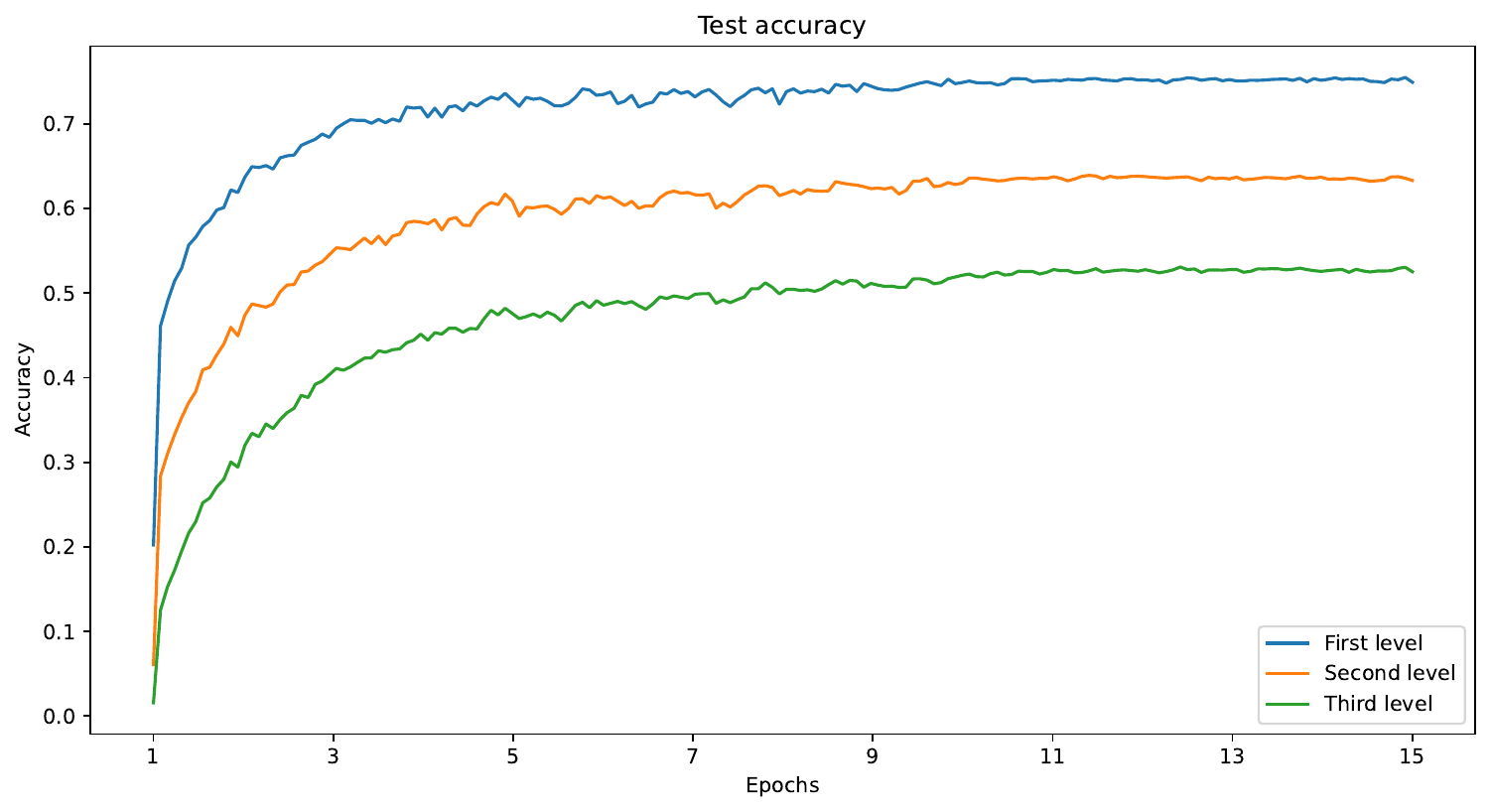}
    \caption{$\ell$-LH-DNN test accuracy.}
    \label{fig:C100_Hl_test}
    \end{subfigure}
    \caption{$\ell$-LH-DNN on CIFAR 100.}
    \label{fig:C100_Hl}
\end{figure}

By looking at the network loss and accuracy during the learning, reported by Figures \ref{fig:C100_B}-\ref{fig:C100_Hl}, one can infer similar considerations to those made for CIFAR10.
Nonetheless, there is another negative phenomenon in B-CNN learning that was less evident in the CIFAR10 study but deserves to be pointed out.
Among the asymptotic losses on the three objectives (see Figure \ref{fig:C100_B}) only the third one converges to zero.
This is a reflection, or the cause depending on the perspective, of the intrinsic incoherence of B-CNN: non-negligible errors on higher levels of the hierarchy are accepted to reach good performance on the lower ones.
It also testifies to the poor use of the potential expressiveness of the network parameters, as LH-DNNs reach comparable performance on the finest categorization with lesser parameters and higher performance on the coarser ones.
On the other hand, LH-DNNs do not (and, by design, cannot) suffer from this phenomenon as shown by Figure \ref{fig:C100_Hs} and \ref{fig:C100_Hl}.

\subsection{Results on Fashion-MNIST}
Table \ref{tab:performance_FM} reports the performance of the two networks using the same metrics as per the two previous benchmarks.
\begin{table}[ht]
    \centering
    \caption{Performance on Fashion-MNIST.}
    \resizebox{\columnwidth}{!}{
    \begin{tabular}{c|c|c|c|c|c|c|}
    \cline{2-7}     & \multicolumn{3}{|c|}{\textbf{Accuracy}} & \multicolumn{3}{|c|}{\textbf{Coherency}} \\
    \cline{2-7} & $\mathbf{1}^{\text{\textbf{st}}}$ \textbf{level} & $\mathbf{2}^{\text{\textbf{nd}}}$ \textbf{level} & $\mathbf{3}^{\text{\textbf{rd}}}$ \textbf{level} & $\mathbf{1}^{\text{\textbf{st}}}$ \textbf{vs} $\mathbf{2}^{\text{\textbf{nd}}}$ & $\mathbf{2}^{\text{\textbf{nd}}}$ \textbf{vs} $\mathbf{3}^{\text{\textbf{rd}}}$ & $\mathbf{1}^{\text{\textbf{st}}}$ \textbf{vs} $\mathbf{3}^{\text{\textbf{rd}}}$ \\
    \hline
    \multicolumn{1}{|c|}{\textbf{B-CNN}} & 99.77\% & 96.31\% & 93.23\% & 99.81\% & 98.39\% & 99.72\% \\
    \hline
    \multicolumn{1}{|c|}{\textbf{$q$-LH-DNN}} & 99.77\% & 96.44\% & 92.68\% & 99.89\% & 99.45\% & 99.85\% \\
    \hline
    \multicolumn{1}{|c|}{\textbf{LH-DNN}} & \cellcolor[HTML]{b7e1cd}99.78\% & \cellcolor[HTML]{b7e1cd}96.69\% & \cellcolor[HTML]{b7e1cd}93.34\% & \cellcolor[HTML]{b7e1cd}99.95\% & \cellcolor[HTML]{b7e1cd}99.62\% & \cellcolor[HTML]{b7e1cd}99.92\% \\
    \hline
    \end{tabular}}
    \label{tab:performance_FM}
\end{table}
Even on this benchmark, the performance of the LH-DNN is superior despite the significantly smaller number of parameters, the reduced training horizon, and the lack of true hyperparameters fine-tuning.
The impact of the proposed technology on hierarchical learning is further emphasized when halving the training time.
In such a small interval, the network (referred to as $q$-LH-DNN in Table \ref{tab:performance_FM}, where $q$ stands for ``quick'') proves to already outperform the B-CNN but on the finest hierarchical level, where the performance is yet comparable.
For the sake of brevity, the plots of the training loss and test accuracy 
are omitted as they do not provide any significant information as to what has been evinced so far for the other two benchmarks.



\subsection{Time and memory consumption}

This subsection is meant to briefly discuss the training time and the networks' memory requirements.
Table \ref{tab:time_C10}, \ref{tab:time_C100}, and \ref{tab:time_FM} report the overall computational time and the number of epochs required for the network training on the three benchmarks.
On the other hand, Table \ref{tab:memory} lists the number of parameters needed by each network to achieve the aforementioned performance.

As expected, LH-DNNs systematically outperform the B-CNNs.
Nonetheless, LH-HDNNs training time is still affected by the time overhead due to the projection blocks initialization.
Indeed, they are recomputed for each batch since they depend on the always-changing parameters of the previous levels.
This fact is testified by a time reduction non-linearly proportional to the ratio of the parameter number.

\begin{table}[ht]
    \centering
    \caption{Time consumption on CIFAR10.}
    \begin{tabular}{|c|c|c|c|}
        \hline
         & \textbf{B-CNN} & \textbf{$s$-HDNN} & \textbf{$\ell$-HDNN}  \\
         \hline
    \textbf{Total time}     & 196.59s & \cellcolor[HTML]{b7e1cd} 74.72s &  83.17s\\
    \hline
    \textbf{Epochs} & 60 & \cellcolor[HTML]{b7e1cd} 20 & \cellcolor[HTML]{b7e1cd} 20\\
    \hline
    \end{tabular}
    \label{tab:time_C10}

\bigskip\bigskip

    \centering
    \caption{Time consumption on CIFAR100.}
    \begin{tabular}{|c|c|c|c|}
        \hline
         & \textbf{B-CNN} & \textbf{$s$-HDNN} & \textbf{$\ell$-HDNN}  \\
         \hline
    \textbf{Total time}     & 225.21s & 76.86s & \cellcolor[HTML]{b7e1cd} 62.80s\\
    \hline
    \textbf{Epochs} & 80 &  20 & \cellcolor[HTML]{b7e1cd} 15\\
    \hline
    \end{tabular}
    \label{tab:time_C100}

\bigskip\bigskip

    \caption{Time consumption on Fashion-MNIST.}
    \begin{tabular}{|c|c|c|c|}
        \hline
         & \textbf{B-CNN} & \textbf{$q$-LH-DNN} & \textbf{LH-DNN}  \\
         \hline
    \textbf{Total time}     & 357.85s & \cellcolor[HTML]{b7e1cd} 34.13s & 87.24s\\
    \hline
    \textbf{Epochs} & 80 & \cellcolor[HTML]{b7e1cd} 9 & 20\\
    \hline
    \end{tabular}
    \label{tab:time_FM}

\bigskip\bigskip

    \caption{Parameters required by the best LH-DNN against the B-CNN on each benchmark.}
    \begin{tabular}{|c|c|c|}
        \hline
         & \textbf{B-CNN} & \textbf{LH-DNN} \\
         \hline
    \textbf{CIFAR10/100}     & 12,382,035 & \cellcolor[HTML]{b7e1cd} 5,746,131 \\
    \hline
    \textbf{Fashion-MNIST} & 38,755,282 & \cellcolor[HTML]{b7e1cd} 2,462,290\\
    \hline
    \end{tabular}
    \label{tab:memory}
\end{table}

\section{Conclusions}
\label{sec:conclusions}
The manuscript tackles the problem of hierarchical classification from a novel perspective mainly based on the use of non-standard analysis (NSA), a branch of mathematical logic dealing with infinite and infinitesimal numbers.
In particular, the hierarchical problem is first reinterpreted as a lexicographic problem over the per-level accuracy, and then it is turned into a non-standard problem by using already existing results in that field.
After a deep mathematical analysis, the work elaborates and proposes for the first time in literature a standard network topology, called LH-DNN, able to implement non-standard learning, which is shown to not be naively implementable without serious drawbacks.
The key element of the new topology is a projection block that guarantees the non-interference of the learning at different levels of the hierarchy and a high degree of coherency between the labels graph structure and the network categorization.
To assess the effectiveness of LH-DNNs on hierarchical problems, three different networks have been tested on CIFAR10, CIFAR100, and Fashion-MNIST against one of the most influential and effective approaches in this domain: the B-CNN.
The results testify that not only LH-DNNs can reach comparable or even better performance, but they do so with a significantly higher coherency with respect to the label hierarchy, with remarkably fewer parameters (from one-sixteenth to one-half), and in a notably shorter amount of time (both in terms of epochs and overall training time).
This means that the topology induced by NSA is able to implement a better use of the network parameters.
Therefore, evidence confirms the theoretical expectations and encourages further studies on LH-DNNs as they may represent a remarkable and effective novelty in the domain.

\section*{Acknowledgments}
\noindent This work has been partially supported by:\\
- the Italian Ministry of Education and Research (MIUR) in the framework of the FoReLab project (Departments of Excellence);\\
- PNRR - M4C2 - Investimento 1.3, Partenariato Esteso PE00000013 - FAIR - ``Future Artificial Intelligence Research'' - Spoke 1 ``Human-centered AI'' (program funded by the European Commission under the NextGeneration EU programme);\\
- University of Pisa Green Data Center (staff members: Dr. M. Davini and Dr F. Pratelli).

\bibliographystyle{IEEEtran}
\bibliography{references}

\onecolumn
\include{supplementary}

\end{document}

%% file: supplementary.tex










\section{Proof of Proposition 3}

The proof requires to verify the satisfaction of (13)-(14) and proceeds by induction.\\
        \textit{Case $n=2$.} Given the pair of parameters $\overline{\theta}_s = (\overline{\theta}_{s_1}, \overline{\theta}_{s_2})$, $z$ is defined as
    \begin{equation*}
        z \coloneqq \rho(\overline{\theta}_{s_1}\rho(\overline{\theta}_{s_2} x)),
    \end{equation*}
    along with the partial quantities
    \begin{equation*}
        k_2 \coloneqq \overline{\theta}_{s_2} x,\quad z_2 \coloneqq \rho(k_2),\quad k_1 \coloneqq \overline{\theta}_{s_1} k_2.
    \end{equation*}
    After the backpropagation of $f_2$ using gradient descent, it turns out that (the point where the gradient is calculated is omitted for the sake of brevity)
    \begin{equation*}
    \begin{split}
        z \coloneqq& \rho((\overline{\theta}_{s_1} - \lambda\nabla_{\theta_{s_1}}f_2) \rho((\overline{\theta}_{s_2}-\lambda\nabla_{\theta_{s_2}}f_2) x)) = \rho((\overline{\theta}_{s_1} - \lambda\nabla_{\theta_{s_1}}f_2) \rho(\overline{\theta}_{s_2}x -\lambda\nabla_{\theta_{s_2}}f_2 x)) \simeq\\
        \simeq& \rho((\overline{\theta}_{s_1} - \lambda\nabla_{\theta_{s_1}}f_2)( \rho(k_2) - \rho_{k_2}' \lambda\nabla_{\theta_{s_2}}f_2x)) = \rho((\overline{\theta}_{s_1} - \lambda\nabla_{\theta_{s_1}}f_2) (z_2 - \rho_{k_2}' \lambda\nabla_{\theta_{s_2}}f_2 x)) = \\
        =& \rho(\overline{\theta}_{s_1}z_2 - \lambda\nabla_{\theta_{s_1}}f_2 z_2 - \overline{\theta}_{s_1}\rho_{k_2}' \lambda\nabla_{\theta_{s_2}}f_2 x +\underbrace{\lambda\nabla_{\theta_{s_1}}f_2\rho_{k_2}' \lambda\nabla_{\theta_{s_2}}f_2 x)}_{2^{\text{nd}} \text{ order term, ignored}} \simeq\\
        \simeq& \rho(\overline{\theta}_{s_1}z_2 - \lambda\nabla_{\theta_{s_1}}f_2 z_2 - \overline{\theta}_{s_1}\rho_{k_2}' \lambda\nabla_{\theta_{s_2}}f_2 x) \simeq \rho(k_1) - \rho_{k_1}'\lambda\nabla_{\theta_{s_1}}f_2 z_2 - \rho_{k_1}' \overline{\theta}_{s_1}\rho_{k_2}' \lambda\nabla_{\theta_{s_2}}f_2 x=\\
        =&z - \rho_{k_1}'\lambda\nabla_{\theta_{s_1}}f_2 z_2 - \rho_{k_1}' \overline{\theta}_{s_1}\rho_{k_2}' \lambda\nabla_{\theta_{s_2}}f_2 x.
    \end{split}
    \end{equation*}
    Thus, Equation (13) imposes
    \begin{equation*}
    \begin{split}
        \langle\overline{\theta}_1,  \rho_{k_1}'\nabla_{\theta_{s_1}}f_2 z_2 + \rho_{k_1}' \overline{\theta}{s_1}\rho_{k_2}' \nabla_{\theta_{s_2}}f_2 x\rangle &= \langle\overline{\theta}_1\rho_{k_1}',  \nabla_{\theta_{s_1}}f_2 z_2 + \overline{\theta}{s_1}\rho_{k_2}' \nabla_{\theta_{s_2}}f_2 x\rangle =\\
        &= \underbrace{\langle\overline{\theta}_1\rho_{k_1}',  \nabla_{\theta_{s_1}}f_2 z_2\rangle}_{=0\,\, \text{by Theorem }5} + \langle\overline{\theta}_1\rho_{k_1}', \overline{\theta}{s_1}\rho_{k_2}' \nabla_{\theta_{s_2}}f_2 x\rangle =\\
        &=\langle\overline{\theta}_1\rho_{k_1}', \overline{\theta}{s_1}\rho_{k_2}' \nabla_{\theta_{s_2}}f_2 x\rangle = 0.
    \end{split}
    \end{equation*}
    By definition,
    \begin{equation*}
        \nabla_{\theta_{s_2}}f_2\Big|_{(P_{\overline{\theta}_1^T\rho_{k_1}'} \overline{\theta}_2, \overline{\theta}_s,x)} = \varepsilon \rho_{k_2}' \overline{\theta}_{s_1}^T \rho_{k_1}' P_{\overline{\theta}_1^T\rho_{k_1}'}\overline{\theta}_2x^T = U P_{\overline{\theta}_1^T\rho_{k_1}'}\overline{\theta}_2x^T
    \end{equation*}
    and so
    \begin{equation*}
        \langle\overline{\theta}_1\rho_{k_1}', \overline{\theta}_{s_1}\rho_{k_2}' \nabla_{\theta_{s_2}}f_2 x\rangle = \underbrace{\overline{\theta}_1^T\rho_{k_1}' \underbrace{\overline{\theta}_{s_1}\rho_{k_2}'U P_{\overline{\theta}_1^T\rho_{k_1}'}}_{\perp\overline{\theta}_1^T\rho_{k_1}'}}_{=0}\overline{\theta}_2x^Tx = 0.
    \end{equation*}
    This guarantees that condition (13) is satisfied.
    To verify the satisfaction of (14), it must hold that
    \begin{equation*}
        \left\langle\nabla_{\theta_{s_2}}f_2 \Big|_{(P_{\overline{\theta}_1^T\rho_{k_1}'}\overline{\theta}_2, \overline{\theta}_s,x)}, \nabla_{\theta_{s_2}}f_2\right\rangle \ge 0,
    \end{equation*}
    that is
    \begin{equation*}
        tr\left(x\overline{\theta}_2^TP_{\overline{\theta}_1^T\rho_{k_1}'}^T\rho_{k_1}'\overline{\theta}_{s_1}^T\rho_{k_2}'\varepsilon^2\rho_{k_2}'\overline{\theta}_{s_1}\rho_{k_1}'\overline{\theta}_2x^T\right) \ge 0.
    \end{equation*}
    For the same reasons as in Theorem 5, such a condition holds.
    This completes the proof for the current case.\\
    \textit{Case $n>2$.} Given the tuple of parameters $\overline{\theta}_s = (\overline{\theta}_{s_1}, \ldots,\overline{\theta}_{s_n})$, let one define
    \begin{equation*}
        k_n \coloneqq \overline{\theta}_{s_n} x,\quad k_i \coloneqq \overline{\theta}_{s_i} z_{i+1}, \quad z_i \coloneqq \rho(k_i),\quad z \coloneqq z_1.
    \end{equation*}
    Assume that the proposition holds for the first $n\um1$ parameters, that is conditions (13)-(14) hold for the gradient (the point at which it is calculated is omitted for brevity)
    \begin{equation*}
        \nabla_{\theta_{s_{1:n-1}}}f_2\Big|_{(P_{\overline{\theta}_1^T\rho_k'}\overline{\theta}_2, \overline{\theta}_s,x)} =
        \begin{bmatrix}
            \nabla_{\theta_{s_1}}f_2\\
            \vdots\\
            \nabla_{\theta_{s_{n-1}}}f_2\\
            0
        \end{bmatrix}.
    \end{equation*}
    Then, one needs to prove that the two also hold for the following gradient
    \begin{equation*}
        \nabla_{\theta_s}f_2\Big|_{(P_{\overline{\theta}_1^T\rho_k'}\overline{\theta}_2, \overline{\theta}_s,x)} =
        \begin{bmatrix}
            \nabla_{\theta_{s_1}}f_2\\
            \vdots\\
            \nabla_{\theta_{s_n}}f_2
        \end{bmatrix}.
    \end{equation*}

    Through a linear approximation, the impact of the gradient on the optimization is
    \begin{equation*}
        k_n \leftarrow k_n - \lambda\nabla_{\theta_{s_n}}f_2 x,\quad z_n \leftarrow z_n - \lambda\rho_{k_n}'\nabla_{\theta_{s_n}}f_2 x,
    \end{equation*}
    and, more in general,
    \begin{equation*}
        z \leftarrow z - \lambda \sum{i=1}^{n-1} \prod{j=1}^{i-1} \left( \rho_{k_j}' \overline{\theta}_{s_j}  \right) \rho_{k_i}' \nabla_{\theta_{s_i}}f_2 z_{i+1} - \lambda \prod{j=1}^{n-1} \left(\rho_{k_j}' \overline{\theta}_{s_j}\right) \rho_{k_n}' \nabla_{\theta_{s_n}}f_2 x.
    \end{equation*}
    As repeatedly done in the previous proofs, condition (13) holds if 
    \begin{equation*}
        \begin{split}
        &\left\langle\overline{\theta}_1, \sum_{i=1}^{n-1} \prod_{j=1}^{i-1} \left( \rho_{k_j}' \overline{\theta}_{s_j}  \right) \rho_{k_i}' \nabla_{\theta_{s_i}}f_2 z_{i+1} + \prod_{j=1}^{n-1} \left(\rho_{k_j}' \overline{\theta}_{s_j}\right) \rho_{k_n}' \nabla_{\theta_{s_n}}f_2 x\right\rangle =\\
        &=\underbrace{\left\langle\overline{\theta}_1, \sum_{i=1}^{n-1} \prod_{j=1}^{i-1} \left( \rho_{k_j}' \overline{\theta}_{s_j}  \right) \rho_{k_i}' \nabla_{\theta_{s_i}}f_2 z_{i+1}\right\rangle}_{=\, 0\;\; \text{by hypothesis of case } n > 2} + \left\langle\overline{\theta}_1,  \prod_{j=1}^{n-1} \left(\rho_{k_j}' \overline{\theta}_{s_j}\right) \rho_{k_n}' \nabla_{\theta_{s_n}}f_2 x\right\rangle = \left\langle\overline{\theta}_1, \prod_{j=1}^{n-1} \left(\rho_{k_j}' \overline{\theta}_{s_j}\right) \rho_{k_n}' \nabla_{\theta_{s_n}}f_2 x\right\rangle = 0.
        \end{split}
    \end{equation*}
    By definition,
    \begin{equation*}
        \nabla_{\theta_{s_n}}f_2 = \varepsilon \rho_{k_n}' \prod_{i=n-1}^{1}\left(\overline{\theta}_{s_i}^T\rho_{k_i}'\right) P_{\overline{\theta}_1^T\rho_{k_1}'} \theta_2x^T,
    \end{equation*}
    implying that the condition can be rewritten as
    \begin{equation*}
    \begin{split}
        &\left\langle\overline{\theta}_1, \prod_{j=1}^{n-1} \left( \rho_{k_j}' \overline{\theta}_{s_j}\right) \rho_{k_n}'\varepsilon^2\rho_{k_n}' \prod_{i=n-1}^{1}\left(\overline{\theta}_{s_i}^T\rho_{k_i}'\right) P_{\overline{\theta}_1^T\rho_{k_1}'} \theta_2x^T x\right\rangle =\\
        &=\left\langle\overline{\theta}_1, \rho_{k_1}'\theta_{s_1}\prod_{j=2}^{n-1} \left( \rho_{k_j}' \overline{\theta}_{s_j}\right) \rho_{k_n}'\varepsilon^2\rho_{k_n}' \prod_{i=n-1}^{1}\left(\overline{\theta}_{s_i}^T\rho_{k_i}'\right) P_{\overline{\theta}_1^T\rho_{k_1}'} \theta_2x^T x\right\rangle =\\
        &=\underbrace{\overline{\theta}_1\rho_{k_1}' \underbrace{\theta_{s_1}\prod_{j=2}^{n-1} \left( \rho_{k_j}' \overline{\theta}_{s_j}\right) \rho_{k_n}'\varepsilon^2\rho_{k_n}' \prod_{i=n-1}^{1}\left(\overline{\theta}_{s_i}^T\rho_{k_i}'\right) P_{\overline{\theta}_1^T\rho_{k_1}'}}_{\perp \overline{\theta}_1\rho_{k_1}'}}_{=0} \theta_2x^T x = 0,
    \end{split}
    \end{equation*}
    verifying its satisfaction.
    On the other hand, condition (14) holds if and only if
    \begin{equation*}
        \left\langle\nabla_{\theta_{s_i}}f_2 \Big|_{(P_{\overline{\theta}_1^T\rho_{k_1}'}\overline{\theta}_2, \overline{\theta}_i,x)}, \nabla_{\theta_{s_i}}f_2\right\rangle \ge 0, \quad \forall\,i=1,\ldots,n.
    \end{equation*}
    By hypothesis, such a condition is already true for $\forall i<n$, and so it reduces to proving that
    \begin{equation*}
        tr\left(x \theta_2^T P_{\overline{\theta}_1^T\rho_{k_1}'}^T \left(\prod_{i=n-1}^{1}\left( \overline{\theta}_{s_i}^T \rho_{k_i}' \right)\right)^T \rho_{k_n}' \varepsilon^2 \rho_{k_n}' \prod_{i=n-1}^{1}\left(\overline{\theta}_{s_i}^T\rho_{k_i}'\right) \theta_2x^T\right) \ge 0,
    \end{equation*}
    which holds for the same considerations as in Theorem 5.